\documentclass[lettersize,journal]{IEEEtran}
\usepackage{amsmath,amsfonts}
\usepackage{amssymb}
\usepackage{algorithmic}
\usepackage{algorithm}
\usepackage{array}
\usepackage[caption=false,font=normalsize,labelfont=sf,textfont=sf]{subfig}
\usepackage{lipsum}
\usepackage{wrapfig}
\usepackage{pifont}
\usepackage{textcomp}
\usepackage{stfloats}
\usepackage{url}
\usepackage{verbatim}
\usepackage{graphicx}
\usepackage{xcolor}
\usepackage{caption}
\usepackage{threeparttable}
\usepackage{multirow}
\usepackage{makecell}
\usepackage{booktabs}
\usepackage[numbers,sort&compress]{natbib}

\begin{document}

\title{Neural Radiance Field in Autonomous Driving: A Survey}

\author{Lei He$^{1}$, Leheng Li$^{2}$, Wenchao Sun$^{1}$, Zeyu Han$^{1}$, Yichen Liu$^{3}$, Sifa Zheng$^{1}$, Jianqiang Wang$^{1}$, Keqiang Li$^{1*}$

\thanks{$^{1}$School of Vehicle and Mobility, Tsinghua University, Beijing, China}
\thanks{$^{2}$ AI Thrust, Information Hub, The Hong Kong University of Science and Technology - Guangzhou Campus, Guangzhou, China}
\thanks{$^{3}$School of Engineering Mathematics and Technology, University of Bristol, Bristol, UK}
\thanks{$^{*}$Correspondence: {\tt likq@tsinghua.edu.cn}}
}



\maketitle

\begin{abstract}
Neural Radiance Field (NeRF) has garnered significant attention from both academia and industry due to its intrinsic advantages, particularly its implicit representation and novel view synthesis capabilities. With the rapid advancements in deep learning, a multitude of methods have emerged to explore the potential applications of NeRF in the domain of Autonomous Driving (AD). However, a conspicuous void is apparent within the current literature. To bridge this gap, this paper conducts a comprehensive survey of NeRF's applications in the context of AD. Our survey is structured to categorize NeRF's applications in Autonomous Driving (AD), specifically encompassing perception, 3D reconstruction, simultaneous localization and mapping (SLAM), and simulation. We delve into in-depth analysis and summarize the findings for each application category, and conclude by providing insights and discussions on future directions in this field. We hope this paper serves as a comprehensive reference for researchers in this domain. To the best of our knowledge, this is the first survey specifically focused on the applications of NeRF in the Autonomous Driving domain.

\end{abstract}

\begin{IEEEkeywords}
Neural Radiance Field,  Autonomous driving, Perception, 3D Reconstruction, SLAM, Simulation
\end{IEEEkeywords}

\section{Introduction}

\IEEEPARstart{N}{eRF}, as an advanced novel view synthesis technology, harnesses the capabilities of volume rendering and implicit neural scene representation to unveil the complexity of 3D scene geometry. It made its debut at ECCV 2020~\cite{mildenhall2020nerf}, rapidly achieving a leading level of visual quality and serving as a wellspring of inspiration for numerous subsequent research endeavors. In recent years, the domain of autonomous driving has made significant strides, with widespread deployment in highway scenarios, although deployment in urban environments is still undergoing rigorous testing. This technological evolution has shifted from its initial reliance on high-precision maps to provide static scene understanding, now emphasizing real-time perception of local environments through bird's-eye view vision. Simultaneously, it has progressed functionally from Level 2 (L2) and is striving towards Level 4 (L4) autonomy. Autonomous driving systems demand a deep understanding of the surrounding environment, encompassing both static scenes and the dynamic interactions among traffic participants, which is a critical prerequisite for effective planning and control. Through self-supervised learning, NeRF has demonstrated its ability to effectively comprehend local scenes, making it an enticing candidate for enhancing autonomous driving capabilities. Over the past two years, NeRF models have found applications in various aspects of autonomous driving, including perception, 3D reconstruction, simultaneously localization and mapping(SLAM), and simulation, as shown in Fig.~\ref{fig:taxonomy}. 

Neural Radiance Fields (NeRF) has emerged as a promising contender in the field of perception, encompassing a range of critical tasks such as object detection, semantic segmentation, and occupancy prediction. The surge in popularity is primarily attributed to its exceptional ability to acquire precise and consistent geometric information. Research in this field can be classified into two main paradigms, differentiated by the utilization of NeRF: "NeRF for data" and "NeRF for model". The former involves the initial training of NeRF, followed by its use to augment the training data of perception tasks. In contrast, the latter adopts a collaborative training strategy for NeRF and perception networks, enabling the perception networks to learn the geometric information captured by NeRF.

In the realm of 3D reconstruction applications, NeRF can be categorized into three primary methods based on the level of scene understanding: dynamic scene reconstruction, surface reconstruction, and inverse rendering. In the first category, dynamic scene reconstruction focus on reconstructing the dynamic scenes with movable agents, mostly with the sequential 3D bounding box annotation and camera parameters. In the second category, surface reconstruction aims to reconstruct explicit 3D surfaces of the scenes, such as mesh. In the third category, inverse rendering aims to disentangle shape, albedo, and visibility from images of driving scenes, to enable applications such as relighting.

As for SLAM applications, the utilization of NeRF can be classified into three primary methods, each geared towards mapping, localization, or a combination of both. As for localization, NeRF is employed to perform real-time image rendering at the current timestamp and estimate the precise pose of the SLAM system by minimizing the reprojection error. While NeRF for mapping primarily focuses on enhancing the mapping capabilities of the SLAM system, which achieves this by incorporating depth maps generated using NeRF, resulting in improved map accuracy. Besides, NeRF is used in some other research to simultaneously enhance the quality of the 3D map and improve the SLAM system's accuracy in pose estimation. These categorizations demonstrate how NeRF can be strategically integrated into a SLAM system to meet specific needs, whether they involve mapping, localization, or a combination of both functionalities. It is worth mentioning that some of the existing NeRF-based SLAM approaches are designed for indoor scenarios, but as the technique is similar to large-scale outdoor environment for autonomous driving, indoor approaches are also reviewed in this paper.

In NeRF simulations, there are two types. The first type divides driving scenes into static and dynamic components, using neural radiance fields for both. It then edits the motion of vehicles or pedestrians to generate new scenes and simulate image data. This type is further split into implicit and explicit approaches, depending on scene representation. The second type centers on simulating LiDAR data from new viewpoints, integrating LiDAR sensing process models with neural radiance fields to depict the scene's geometry. This type is divided into ray and beam models, based on the modeling differences of the LiDAR sensing process.

As remarkable advancements in both academic and industrial domains have unfolded in this field, we present a comprehensive review of recent developments to catalyze further research. The primary contributions of this work can be summarized as follows:

\begin{figure*}[!htb]
    \centering{\includegraphics[width = 0.96\linewidth]{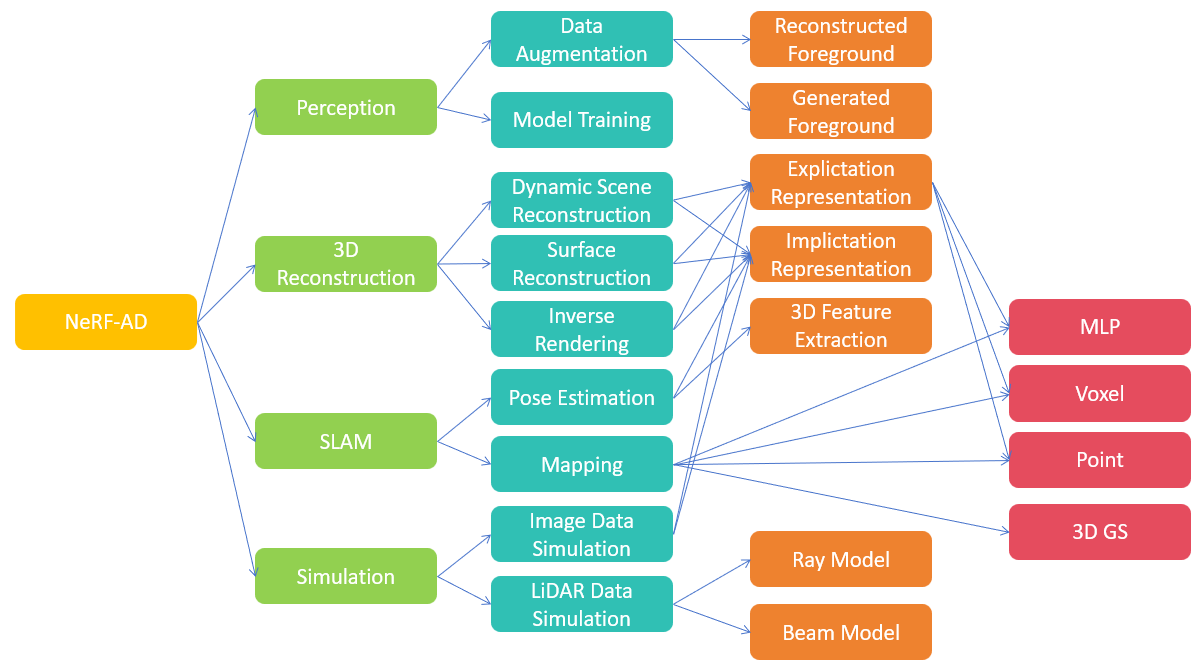}}
    \caption{A taxonomy of Neural Radiance Field in Autonomous Driving.}
    \label{fig:taxonomy}
\end{figure*}

\begin{itemize}
\item{To the best of our knowledge, this is the first comprehensive survey reviewing NeRF's applications in addressing fundamental techniques within the realm of autonomous driving.}
\item{We provide the latest NeRF-AD methodologies, systematically categorizing them based on their core principles and downstream applications.}
\item{We present a comprehensive discussion of NeRF-AD, offering insights into critical research gaps and suggestions for future research directions. }
\end{itemize}

The structure of this paper is outlined as follows. Section~\ref{sec:nerf_prelimi} provides an introduction to the basic principles and background of NeRF. Section~\ref{sec:perception} delves into the analysis of NeRF applications in perception. Section~\ref{sec:reconstruct} conducts an in-depth comparative analysis of NeRF applications in 3D reconstruction. Section~\ref{sec:slam} offers a detailed analysis of NeRF applications in SLAM. Section~\ref{sec:simulation} provides an in-depth analysis of NeRF applications in simulation. Section~\ref{sec:discuss} discusses and predicts future research directions. Finally, Section~\ref{sec:conclu} summarizes the paper.

\section{Neural Radiance Field}\label{sec:nerf_prelimi}

Neural Radiance Fields, first introduced by Mildenhall et al. \cite{mildenhall2020nerf} in 2020, achieved highly realistic view synthesis of complex scenes using only 2D posed images for supervision. NeRF conceptualizes a continuous scene as a 5D vector-valued function. This 5D scene representation, facilitated through an MLP network, is denoted as:  \begin{equation} \label{eq:nerf}
    F(\mathbf{x},\theta,\phi) \xrightarrow{} (\mathbf{c},\sigma)
\end{equation}
where \(\mathbf{x} = (x,y,z)\) denotes the coordinates of points within the scene, \((\theta, \phi)\) refer to the azimuthal and polar viewing angles respectively, \(c = (r,g,b)\) represents the color, and \(\sigma\) signifies the volume density. In practical implementations, \((\theta, \phi)\) are represented as \(\mathbf{d}=(d_x,d_y,d_z)\), a 3D Cartesian unit vector. The network architecture is structured in two stages, where the first stage inputs \(\mathbf{x}\) and outputs \(\sigma\) along with a feature vector. In the second stage, this feature vector is concatenated with the viewing direction \(\mathbf{d}\) to produce the color \(\mathbf{c}\) at that viewpoint. This design enables the network to learn a view-independent \(\sigma\) that relies solely on the in-scene coordinates and a color \(\mathbf{c}\) that depends on both the viewing direction and in-scene coordinates.

\begin{figure*}[!htb]
    \centering{\includegraphics[width = 0.8\linewidth]{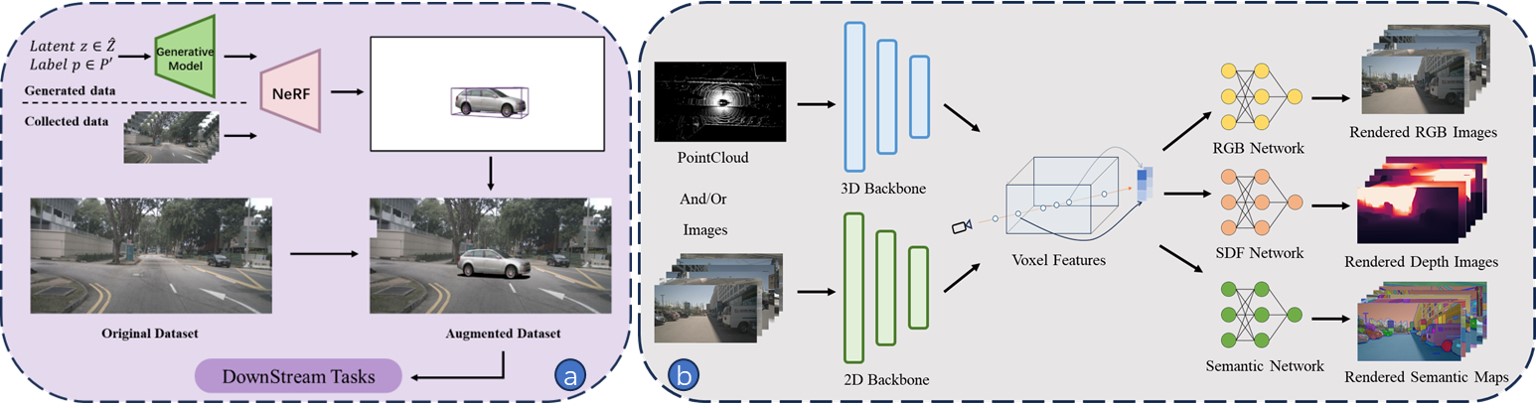}}
    \caption{Overview of NeRF's application in autonomous driving perception: (a) NeRF can be used for data augmentation by reconstructing scenes from either generated data or collected real data. (b) NeRF's implicit representation and neural rendering can be integrated into model training to enhance performance.}
    \label{fig:nerf_perception}
\end{figure*}

Given the volume density and color for each point in a scene, NeRF employs volume rendering, as described in \cite{NERF1989_Raytracing}, to compute the color \(C(\mathbf{r})\) of any camera ray \(\mathbf{r}(t) = \mathbf{o} + t\mathbf{d}\), where \(\mathbf{o}\) is the camera position and \(\mathbf{d}\) is the viewing direction, using the following equation:
\begin{equation} \label{eq:ray}
    C(\mathbf{r}) = \int^{t_2}_{t_1} T(t) \cdot \sigma(\mathbf{r}(t)) \cdot \mathbf{c}(\mathbf{r}(t),\mathbf{d}) \cdot dt,
\end{equation}
where \(\sigma(\mathbf{r}(t))\) and \(\mathbf{c}(\mathbf{r}(t), \mathbf{d})\) correspond to the volume density and color at point \(\mathbf{r}(t)\) on the path of the camera ray oriented in direction \(\mathbf{d}\), and \(dt\) indicates the incremental distance that the ray traverses at each step of the integration.

\(T(t)\) represents the accumulated transmittance, indicating the probability that the ray travels from \(t_1\) to \(t\) without being obstructed. This is defined as follows:
\begin{equation} \label{eq:transmisivity}
    T(t) = \exp(-\int^{t}_{t_1}  \sigma(\mathbf{r}(u))\cdot du).
\end{equation}

Novel views are synthesized by projecting camera rays \(C(\mathbf{r})\) through each pixel of the target image, with the integral evaluated numerically. The original implementation and most subsequent works have employed a stochastic stratified sampling strategy, segmenting the ray into \(N\) uniform segments and selecting a random sample within each. Consequently, the rendering equation is approximated as:
\begin{equation} \label{eq:ray_discrete}
    \hat{C}(\textbf{r}) = \sum_{i=1}^N \alpha_i T_i \textbf{c}_i, \; \text{where} \; T_i = \exp(-\sum_{j=1}^{i-1} \sigma_j \delta_j)
\end{equation}
where \(\delta_i\) denotes the distance between the \(i\)th and \(i+1\)th sample points. \((\sigma_i, c_i)\) are the density and color at the \(i\)th sample point along the given ray, as computed by the MLP. \(\alpha_i = 1 - \exp(-\sigma_i \delta_i)\) represents the opacity at the \(i\)th sample point.

To capture finer details in models, the NeRF framework often incorporates positional encoding. In the original paper, this technique, denoted as \(\gamma\), is applied to each component of the scene coordinate \(\mathbf{x}\) (which is normalized to the range \([-1,1]\)) and the viewing direction \(\mathbf{d}\). This encoding enhances the model's ability to represent high-frequency functions.

For each pixel, the MLP parameters are optimized using a square error photometric loss. Across the entire image, this loss is calculated as follows:
\begin{equation}\label{eq:photo-loss}
   L = \sum_{r \in R} || \hat{C}(\textbf{r}) - C_{gt}(\textbf{r})||_2^2
\end{equation}
where \(C_{gt}(\mathbf{r})\) represents the actual color of the pixel in the training image corresponding to ray \(\mathbf{r}\), and \(R\) denotes the collection of rays linked to the image being synthesized.

To capture finer details in models, the NeRF framework often incorporates positional encoding. In the original paper, this technique, denoted as \(\gamma\), is applied to each component of the scene coordinate \(\mathbf{x}\) (which is normalized to the range \([-1,1]\)) and the viewing direction \(\mathbf{d}\). This encoding enhances the model's ability to represent high-frequency functions. This encoding is calculated as follows:
\begin{multline}
    \gamma(v) = (sin(2^0 \pi v), cos(2^0 \pi v), sin(2^1 \pi v),  cos(2^1 \pi v), \\
    ..., sin(2^{N-1} \pi v), cos(2^{N-1} \pi v)),
\end{multline}

In the original paper, the positional encoding levels were set at \(N=10\) for \(\mathbf{x}\) and \(N=4\) for \(\mathbf{d}\), optimizing the model's sensitivity to spatial and directional variations.

\section{Perception}\label{sec:perception}
        









%

NeRF demonstrate significant potential in autonomous driving perception tasks, which are categorized into two branches: data augmentation and model training. Data augmentation entails utilizing NeRF's innovative view synthesis capabilities to conduct photorealistic data augmentation for training datasets, while model training involves integrating neural rendering into the training process to capture geometric details and enhance performance. This paper delineates the pipelines of these two branches, as illustrated in Fig. \ref{fig:nerf_perception}.

\subsection{Data Augmentation}
Driving scenes are widely recognized for their remarkable diversity and complexity, making it infeasible to capture all scenarios due to the long-tail problem and high costs. Data augmentation stands as an effective technique to enrich training datasets and enhance model performance. Various studies\cite{abu2018augmented, cabon2020virtual, gaidon2016virtual, richter2016playing} utilize graphic engines to synthesize training data, thereby introducing a sim-to-real domain gap. NeRF, however, exhibits a smaller domain gap as it is trained to approximate realistic images.

\begin{figure}[htbp]
    \centering{\includegraphics[width = \linewidth]{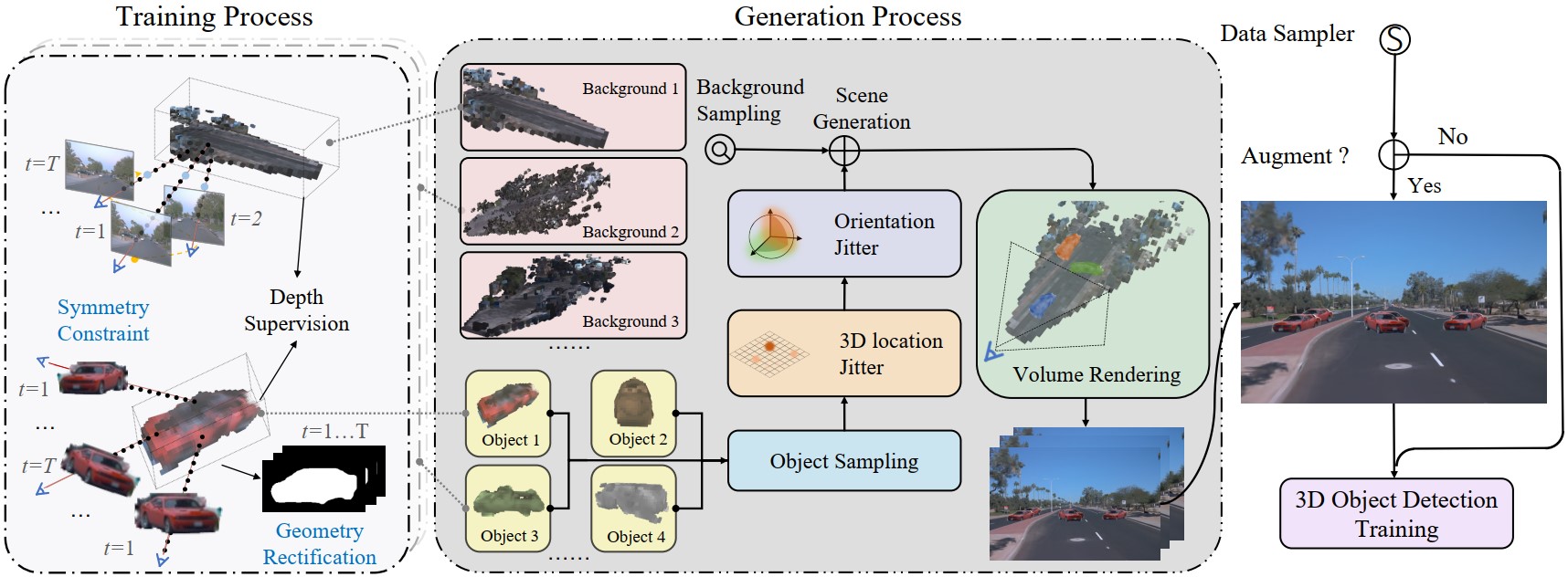}}
    \caption{The pipeline of Drive-3DAug\cite{3daug}.}
    \label{fig:3daug}
\end{figure}

Drive-3DAug\cite{3daug} pioneered research in 3D data augmentation for camera-based 3D perception and demonstrated that NeRF is an effective solution for this purpose. Unlike traditional 2D image augmentation techniques, which are limited to operations on the image plane, such as rotation and copy-and-paste, 3D augmentation has the potential to significantly improve model performance, which has been witnessed in LiDAR-based 3D perception tasks. As shown in Fig. \ref{fig:3daug}, Drive-3DAug comprises two stages: the initial training stage decomposes scenes into background and foreground and constructs 3D models using NeRF, which then serve as reusable digital assets. The subsequent stage involves combining the background with manipulated foreground to create new driving scenes and utilizes volume rendering to produce augmented images. Through 3D data augmentation, object detection models trained with NeRF-based augmentation exhibit superior performance compared to those trained with only 2D data augmentation.

Neural Radiance Fields (NeRF) are leveraged to reconstruct scenes using not only collected sensor data but also label-aware data synthesized by generative models, with the specific aim of reducing annotation costs. Lift3D\cite{lift3d} explores the combination of Generative Adversarial Networks (GAN) and Neural Radiance Fields (NeRF) with the aim of generating data for 3D perception tasks. Initially, pretrained StyleGAN2 is utilized to densely sample images with pose labels. It is assumed that the first 8 layers of the latent code control poses, while the remaining layers influence shape and appearance. A 3D car model from ShapeNet is used to obtain rendered car images from different viewpoints and their corresponding pose labels. Subsequently, an optimization-based GAN inversion method is employed to find the corresponding template latents of the first 8 layers, associating these template latent layers with meaningful pose information. The latent-pose pairs are then incorporated into a 3D shared conditional NeRF, following a 2D-to-3D pipeline. This process eliminates the need for a 2D upsampler, as required by previous methods, and enables the synthesis of images in any resolution. Finally, the trained NeRF can be used to render augmented images for downstream task training.


Based on Lift3D\cite{lift3d}, Adv3D\cite{li2023adv3d} present an innovative exploration of modeling adversarial examples within the context of NeRF, integrating primitive-aware sampling and semantic-guided regularization for 3D patch attacks with camouflage adversarial texture. Their approach involves training an adversarial NeRF to minimize the confidence of 3D detectors for surrounding objects in the training set, resulting in strong generalization capabilities across various poses, scenes, and 3D detectors. Additionally, the paper introduces a defense mechanism against these attacks, employing adversarial training through data augmentation. The intersection of adversarial examples and 3D modeling showcased in this work indicates potential implications for the security and robustness of 3D perception systems, offering valuable insights for applications including autonomous vehicles, robotics, and augmented reality.

\subsection{Model Training}
Several studies have investigated the use of NeRF for data augmentation, but there is a growing body of research that integrates NeRF representation into models to enhance performance. By harnessing implicit scene representation and neural rendering, NeRF effectively bridges the gap between 3D scenes and 2D images, makinging it suitable for a variety of 3D perception tasks.

NeRF has exhibited remarkable performance in scene reconstruction and consequently has found natural applications in perception tasks related to scene completion. BTS\cite{wimbauer2023behind} were among the first to apply volume rendering in single-view reconstruction. Their approach involves inferring an implicit density field as a meaningful geometric scene representation instead of relying solely on depth prediction, which can only reason about visible areas in the image. They utilize an encoder-decoder network to predict a pixel-aligned feature map from the input image. To compute the density value at a given 3D point, features are bilinearly sampled from the feature maps after the 3D point is projected onto the image. Subsequently, along with the features, the depth value of this point and positional encoding are input into a multi-layer perceptron (MLP) to predict the density. The depth can be generated as a by-product of the density field, and for novel view synthesis, colors are sampled from other views rather than being predicted by the MLP, thus drastically reducing the complexity of the distribution along a ray, since density distributions tend to be simple. Multiple views, apart from the input view, are utilized in the training process. These views are divided into two sets, namely $N_{loss}$ and $N_{render}$. Colors are sampled from $N_{render}$ and then used to reconstruct $N_{loss}$, where the photometric consistency between the reconstructed view and $N_{loss}$ serves as the training signal for the density field. This training strategy facilitates the ability to reason about occluded areas in the input view, provided they are visible to other views.

\begin{figure}[htbp]
    \centering{\includegraphics[width = \linewidth]{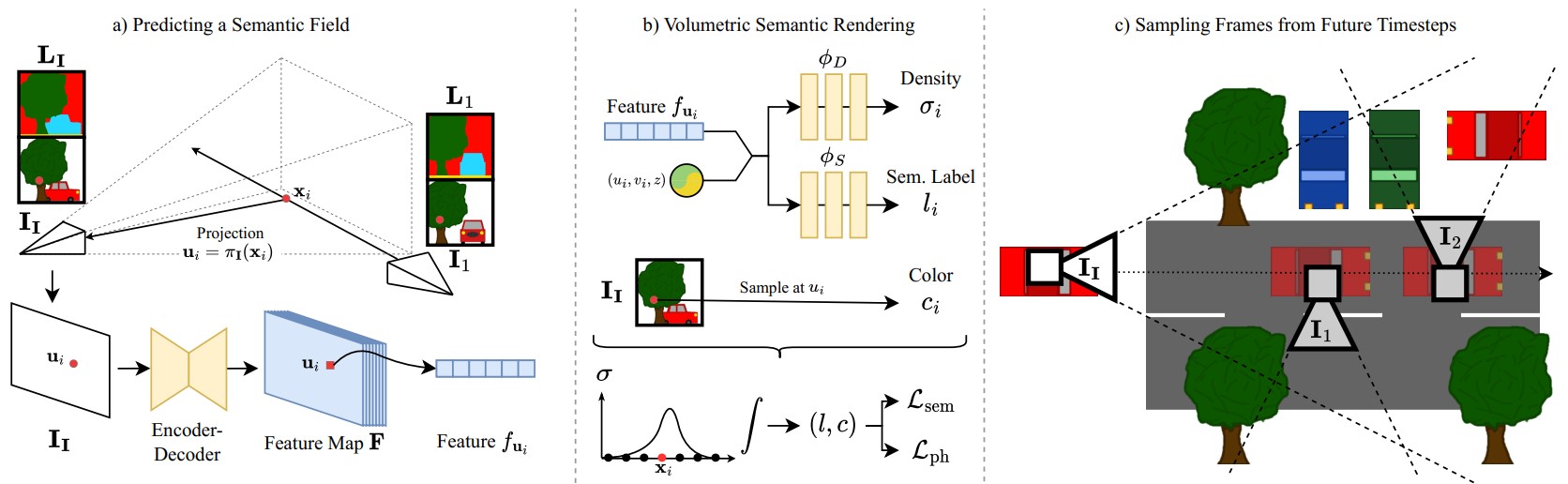}}
    \caption{The pipeline of S4C\cite{hayler2023s4c}.}
    \label{fig:s4c}
\end{figure}

The reasoning of occluded areas is highly relevant to semantic scene completion, as investigated in the work S4C\cite{hayler2023s4c}. As shown in Fig. \ref{fig:s4c}, the processing pipeline is based on BTS\cite{wimbauer2023behind} but incorporates a semantic field in parallel with the density field, enabling the rendering of a semantic map. The disparity between the semantic map and the pseudo-ground truth labels obtained from an off-the-shelf segmentation network provides an additional training signal. Since supervision provided from a single viewpoint only offers training signals for observed areas, it is crucial to strategically select training views. Therefore, sideways-facing views with a random offset from the input view are chosen for training, thereby enhancing diversity and improving the quality of predictions, particularly for further away regions.

Being capable of capturing accurate geometry, NeRF can also be applied to occupancy prediction tasks. SimpleOccupancy\cite{gan2023simple} made an attempt at 3D occupancy estimation, focusing solely on geometry estimation and setting it apart from other similar works. They utilize a shared backbone to extract image features and then apply bilinear interpolation in a parameter-free manner to lift these features to 3D volume space. A 3D convolution network and position embedding are employed for 3D volume feature aggregation. Subsequently, the occupancy probability value is obtained by applying a Sigmoid function. The training process can be supervised using two manners: one involves directly computing a classification loss based on the occupancy probability, while the other utilizes volume rendering to obtain a depth map and supervises it against depth labels. The results indicate that depth loss outperforms classification loss across various metrics, demonstrating the effectiveness of volume rendering.

\begin{figure}[htbp]
    \centering{\includegraphics[width = \linewidth]{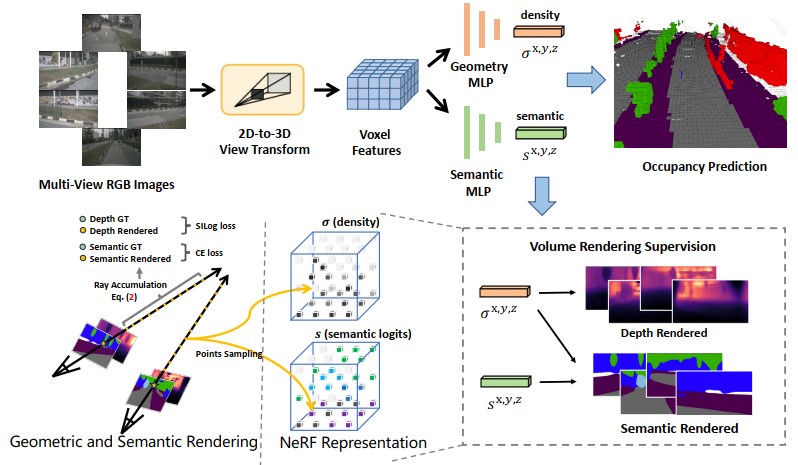}}
    \caption{The pipeline of UniOcc\cite{pan2023uniocc}.}
    \label{fig:uniocc}
\end{figure}

UniOcc\cite{pan2023uniocc} utilizes volume rendering to integrate 2D and 3D representation supervision. Similar to previous research, the approach involves the use of a 2D image encoder, 2D-3D view transformer, and a 3D encoder to generate 3D voxel features, as depicted in Fig. \ref{fig:uniocc}. However, unlike existing methods, UniOcc converts occupancy into NeRF-style representation instead of directly employing an occupancy head for occupancy estimation. It achieves this by using two separate MLPs to predict the density and semantic logits of the voxels. Subsequently, geometric and semantic rendering techniques are applied based on the density and semantic logits to generate 2D depth and semantic logits, which can be supervised by 2D labels. Given the sparsity of viewpoints, temporal frames are introduced as supplementary perspectives after filtering moving objects by semantic categories. As a result of the architectural design and various optimization techniques, UniOcc achieved a 3rd place ranking in the CVPR 2023 3D Occupancy Prediction Challenge.

RenderOcc\cite{pan2023renderocc} demonstrated that 3D occupancy labels are not only expensive, but may also impede model performance due to the inherent ambiguity of occupancy annotation. This limitation constrains the usability and scalability of 3D occupancy models. As a result, they made the pioneering attempt to train 3D occupancy networks solely using 2D labels and achieved competitive performance compared to those supervised by 3D labels. The model architecture shares similarities with UniOcc\cite{pan2023uniocc} but takes a step forward by incorporating temporal frames. Auxiliary rays from adjacent frames are introduced to reinforce multi-view consistency, albeit at the expense of high memory and computational costs, necessitating ray sampling during training. Random sampling may lead to the discarding of many valuable rays, and rays from adjacent frames may cause misalignment due to the movement of dynamic objects. To address these issues simultaneously, a weighted ray sampling method is proposed. This method assigns lower probability weights in random sampling to rays associated with low-information-density backgrounds or dynamic objects, thereby reducing the likelihood of their being sampled. Consequently, this increases information density and mitigates temporal misalignment.

The NeRF model is also applied to the 3D detection task in MonoNeRD\cite{xu2023mononerd}. Utilizing scene geometry to improve the detector's performance is a common approach, and depth estimation has been widely adopted in previous work. However, this often results in sparsity in the 3D representations and significant information loss. In MonoNeRD, a NeRF-like representation is employed for dense 3D geometry. Initially, a camera frustum with multiple depth planes is constructed to extract image features in a Query-Key-Value manner. Subsequently, two convolution blocks transform the frustum features into SDF and RGB features, where the SDF features can be further converted to density features. These density and RGB features are utilized for volume rendering to supervise the model using depth loss and RGB loss. As the irregular frustum features cannot be directly used by downstream detection modules, 3D voxel features are constructed by trilinear sampling from the frustum features and then fed to the detection head. It is also noted that other views can be utilized for rendering as long as their frustum overlaps with the original one.

NeRF is particularly well-suited for static perception tasks, such as map construction, due to its inherent property of multi-view consistency. In contrast to current on-board map construction approaches, MV-Map\cite{xie2023mv} focuses on off-board HD-Map generation. The methodology involves using a pre-trained bird's-eye view (BEV) segmentation model to generate BEV features and semantic maps for each frame in a frame-centric manner. These are then aggregated globally in a region-centric manner. The BEV features serve as input to an uncertainty network, which generates confidence maps. The semantic content of a grid is determined by a weighted average of all semantic maps that overlap with it. A voxelized NeRF is employed to cover the entire scene and capture consistent multi-view geometry. Additionally, the study suggests that the predicted semantics at a position are more reliable when they reside to object surfaces. So for each voxel, the 3D position of its projected pixel location can be reconstructed by the trained NeRF, and the residual between the voxel center and the 3D position is used as an augmented input for the uncertainty network, representing the proximity of the voxel to object surfaces.

\begin{figure}[htbp]
    \centering{\includegraphics[width = \linewidth]{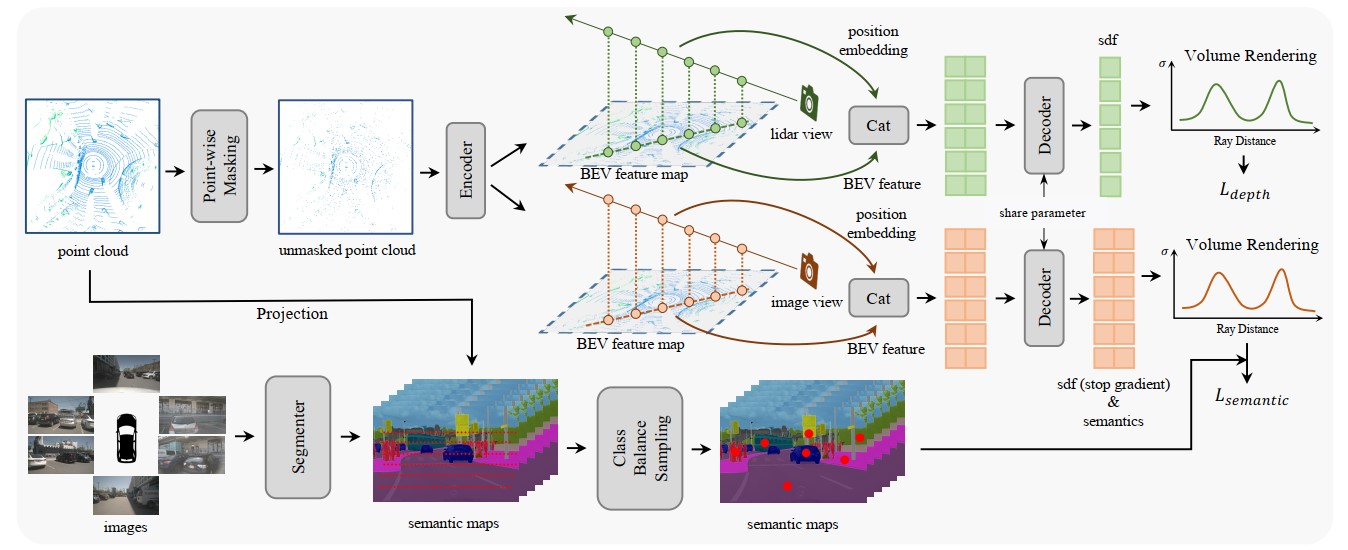}}
    \caption{The overall pipeline of PRED\cite{yang2023pred}.}
    \label{fig:pred}
\end{figure}

Beyond specific image-based tasks, volume rendering has the potential to bridge the gap between point clouds and images, facilitating representation learning for pre-training. In their work, PRED\cite{yang2023pred} focuses on the pre-training of LiDAR point clouds. As depicted in Fig. \ref{fig:pred}, the authors first apply point-wise masking to the input point cloud, preserving the semantics of objects even in sparser regions. The remaining point cloud is then transformed into a Bird's Eye View (BEV) feature map by an encoder, which is subsequently mapped to Signed Distance Function (SDF) and semantic information by a decoder. Due to the absence of color information in the point cloud, only semantic and depth supervision are used after volume rendering. By leveraging semantic rendering, the comprehensive information and rich semantics of images enhance the performance of point cloud pre-training across various tasks.

UniPAD\cite{unipad}, going a step further, propose a flexible pre-training method that enables seamless integration into both 2D and 3D frameworks. This method comprises two components: a modality-specific encoder and a volumetric rendering decoder. For point cloud data, a 3D backbone is utilized for feature extraction, while for multi-view image data, a 2D backbone is employed to extract image features, which are subsequently transformed into a 3D voxel representation. Following the approach of MAE\cite{He_2022_CVPR}, a masking strategy is employed to input data to effectively learn representations. The voxel features are then converted to signed distance function (SDF) value and color value. By integrating predicted colors and sampled depth along rays, images and depth maps are rendered and supervised by groundtruth. In order to reduce memory costs, depth-aware ray sampling is introduced to sample only rays within a depth threshold, thus disregarding distant background.

\section{3D Reconstruction}\label{sec:reconstruct}

As shown in Tab.~\ref{Tab:reconstruction}, we categorized 3D reconstruction into three sub-problems: dynamic scene reconstruction, surface reconstruction, and inverse rendering. We discuss them in the following parts.


\begin{table*}[!htbp]
    \centering
    \caption{Taxonomy of NeRF Reconstruction Research}
    \resizebox{\linewidth}{!}{
    \begin{tabular}{cccc} 
        \toprule
        Categories &\makecell{Representative Research} &Features & Primitives \\
        \midrule
        \multirow{5}{*}{\makecell{Dynamic Scene \\ Reconstruction}}  
        
        & NSG~\cite{ost2021neural} & Use 3D Box to separate objects and background  & MLP \\ 

        & Block-NeRF~\cite{tancik2022block} & Use latent embedding to control illumination & MLP \\ 

        & SUDS~\cite{turki2023suds} &Dynamic hash table & Hash grid \\ 
        
        & EmerNeRF~\cite{yang2023emernerf} & Self-supervised scene flow estimation & Hash grid \\ 
        
        & PVG~\cite{chen2023periodic}, DrivingGaussian~\cite{zhou2023drivinggaussian} & Dynamic 3D Gaussian Splatting & 3D GS \\ 
        \midrule
        
        \multirow{3}{*}{\makecell{Surface \\ Reconstruction}} 
        
        & StreetSurf~\cite{guo2023streetsurf} & Multi-scale hash grid & Hash grid \\ 
        
        & FEGR~\cite{wang2023neural}  & Hybrid 3D representation & Hash grid, Mesh \\ 
        
        & DNMP~\cite{lu2023urban} &Novel scene representations & Hash grid \\ 
        \midrule

        \multirow{3}{*}{\makecell{Inverse \\ Rendering}} 
        
        & FEGR~\cite{wang2023neural}  & Compute illumination via Monte Carlo ray tracing & Hash grid, Mesh \\ 
        
        & UrbanIR~\cite{lin2023urbanir} & Novel scene representations & MLP \\ 
        
        & LightSim~\cite{pun2023lightsim} & Physically-based rendering + learnable deferred rendering & Mesh, MLP \\  

        \bottomrule
    \end{tabular}
    \label{Tab:reconstruction}}
\end{table*}

\begin{figure*}[!t]
    \centering{\includegraphics[width = \linewidth]{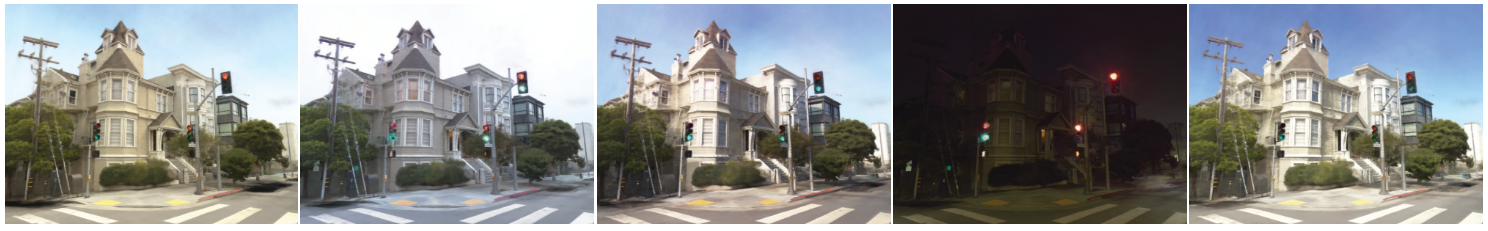}}
    \caption{By learning a latent embedding of each pass of the same place, Block-NeRF~\cite{tancik2022block} can control the lighting effect of rendered images by changing the latent embedding.}
    \label{fig:blocknerf}
\end{figure*}

\begin{figure}[!b]
    \centering{\includegraphics[width = \linewidth]{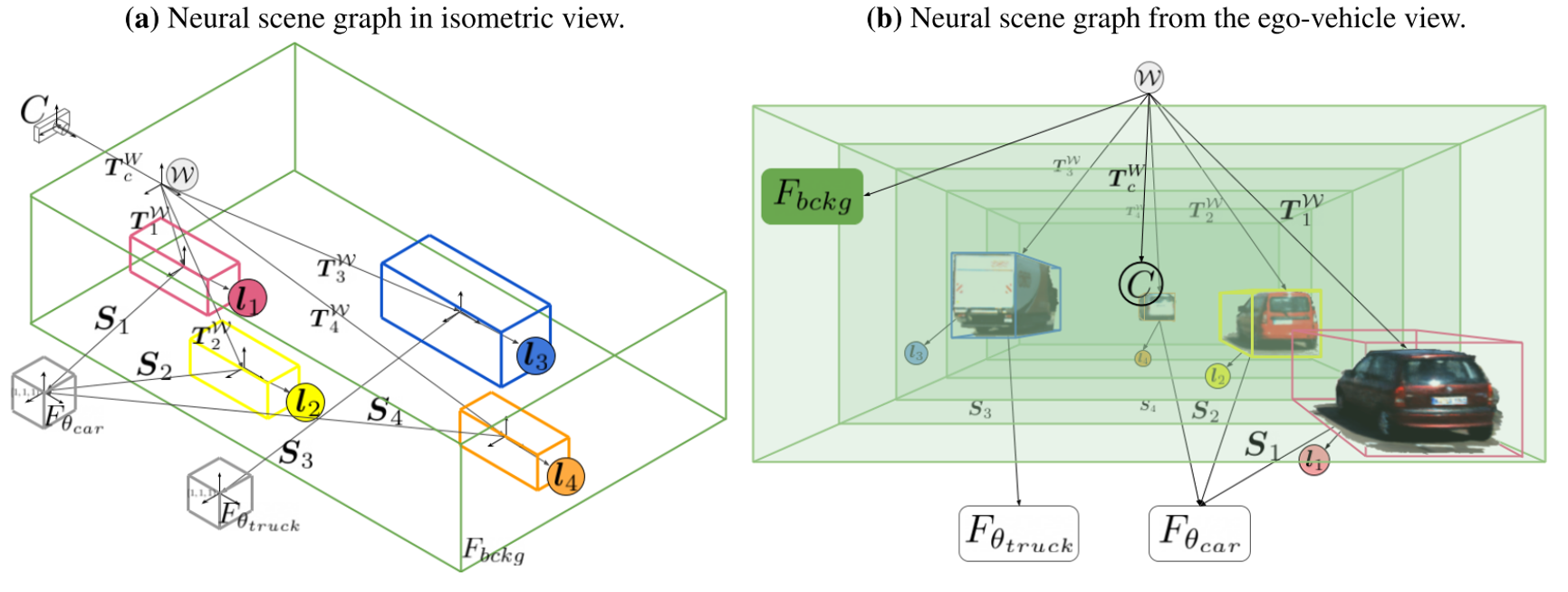}}
    \caption{The pipeline of NSG~\cite{ost2021neural}, which use disentangled 3D graph to represent the objects and scene.}
    \label{fig:NSG}
\end{figure}

\subsubsection{Dynamic Scene Reconstruction}

Neural Scene Graphs (NSG)~\cite{ost2021neural}, for the first time, proposes to reconstruct the 3D dynamic scene with neural scene graphs that are connected by the transformation matrixes. Each node is categorized by dynamic node and static node. The dynamic node can be any dynamic actor (car and pedestrian) that is noted by 3D bounding box in each time stamp. The static node is formulated as a static background. Each node is represented by category-shared MLP and learnable embedding for each instance. During ray casting, NSG first disentangles each node by doing an AABB-ray intersection algorithm~\cite{majercik2018ray} in the 3D box of each node, then asks MLP to process each normalized coordinate. Finally, volume rendering is conducted in a compositional manner of depth order. These scene graph representations enable actor insertion, modification, removal, and rendering from a new viewpoint, all in a unified manner.

To achieve 3D reconstruction of large-scale driving scenes, the Block-NeRF~\cite{tancik2022block} approach utilizes a divide-and-conquer strategy, breaking down the entire scene into individual blocks. Each block is then represented by a specific MLP network. During training, dynamic objects such as cars and pedestrians are masked out using semantic segmentation. To fully harness the multi-pass data collection in the same location, Block-NeRF learns appearance codes similar to NeRF-W~\cite{martin2021nerf} in order to control the lighting and weather of the rendered images. During inference, Block-NeRF is able to generate diverse lighting effects for the same area in the rendered images.


With the LiDAR enhancement, Neural Point Light Fields~\cite{ost2022neural} use LiDAR point clouds as initialization and learn a light field to reconstruct a driving scene. When conducting volume rendering, the method selects a set of K nearest points from a point cloud for each ray. It then utilizes a light field function to predict the color of the ray, taking into account the ray's direction and features aggregated through a multi-head attention module from the closest points.

Similarly, READ~\cite{li2023read} learns a point cloud renderer to reconstruct a 3D scene. The point cloud of the scene is obtained through the matching feature points and dense construction. Then READ learns a neural renderer using a U-net-like network. DGNR~\cite{li2023dgnr} also leverages point clouds as the primitives of 3D representation.

To incorporate map information, MapNeRF~\cite{wu2023mapnerf} proposed a new method to enhance out-of-trajectory driving view synthesis by incorporating map priors in driving scenes into neural radiance fields. Neural Radiance Fields with LiDAR Maps~\cite{chang2023neural} proposes to incorporate LiDAR point cloud prior and GAN to benefit the training of neural radiance field.

To reconstruct large scale scenes, SUDS~\cite{turki2023suds} factorizes the scene into three separate data structures to efficiently encode static, dynamic, and far-field radiance fields. They use a hash grid from instant-ngp as data structures to speed up training and inference. The dynamic branch makes use of 4D spacetime input position and time to index the feature from the hash table. They also use unlabeled inputs including images, point clouds, self-supervised 2D descriptors, and 2D optical flow to learn scene flow and semantic predictions, enabling category- and object-level scene manipulation.

Without relying on ground truth 3D box or pretrained model of depth estimation and optical flow, EmerNeRF~\cite{yang2023emernerf} learns dynamic fields in a self-surprised manner. It first learns a flow field that makes a forward and backward warping into the next or previous frame, then aggregates the per-point features. To enhance the utility for semantic scene comprehension, EmerNeRF proposes to incorporate 2D foundation model features such as DINOv2~\cite{oquab2023dinov2} feature to benefit the training of NeRF. 

In another line, UC-NeRF~\cite{cheng2023uc} trains NeRF in an under-calibrated camera setting. They propose 1) a layer-based color correction to address color inconsistencies in the training images, 2) virtual warping to generate more viewpoint-diverse but color-consistent virtual views for color correction and 3D recovery, and 3) a spatiotemporally constrained pose refinement designed for more robust and accurate pose calibration in multi-camera systems.

\subsubsection{Surface Reconstruction}

FEGR~\cite{wang2023neural} learns to intrinsically decompose the driving scene using a hybrid representation of the 3D scene. Given posed images, FEGR first learns an explicit mesh using a hash grid, then estimates the spatially varying materials and HDR lighting of the underlying scene through their proposed hybrid deferred rendering pipeline. They display satisfactory results on downstream applications such as relighting and virtual object insertion.

StreetSurf~\cite{guo2023streetsurf} develops a multi-view implicit surface reconstruction method for street view using hash tables. They disentangle the large-scale and multi-scale driving scene into three distinct parts based on their distance from the cameras: close-range, distant-view, and sky parts. For each part, they have utilized different models - a cuboid NeuS model for the close-range scene, a hyper-cuboid NeRF++ model for the distant view, and a directional MLP for the sky. Additionally, they have incorporated monocular estimated depth and normal to provide further supervision for the reconstruction process.

To further extract detailed geometric, DNMP~\cite{lu2023urban} proposes to parameterize the entire scene with mesh primitives. The entire scene is voxelized and each voxel is assigned a network to parameterize the geometry and radiance of the local area. The shape of DNMP is decoded from a pretrained latent space to constrain the degree of freedom for robust shape optimization. The radiance features are associated with each mesh vertex of DNMPs for radiance information encoding. 

\subsubsection{Inverse Rendering}

UrbanIR~\cite{lin2023urbanir} learning to infer shape, albedo, and visibility from a single video of driving scenes. It proposes a visibility loss function, which facilitates highly accurate shadow volume estimates within the original scene. This allows for precise editing control, ultimately providing photorealistic renderings of relit scenes and seamlessly inserted objects from any viewpoint.

LightSim~\cite{pun2023lightsim} is a neural lighting camera simulation system that enables diverse, realistic, and controllable data generation. LightSim first builds lighting-aware digital twins at scale from sensor data and decomposes the scene into dynamic actors and static backgrounds with accurate geometry, appearance, and estimated scene lighting. Then LightSim combines physically-based and learnable deferred rendering to perform realistic relighting of modified scenes, such as altering the sun location and modifying the shadows or changing the sun brightness, producing spatially- and temporally-consistent camera videos. 

\subsubsection{Others}

MINE~\cite{li2021mine} learns a generalizable multi-plane image feature grid for Novel View Synthesis.

PVG~\cite{chen2023periodic}, DrivingGaussian~\cite{zhou2023drivinggaussian} and Street Gaussians~\cite{yan2024street} use 3D Gaussian Splatting~\cite{kerbl20233d} to reconstruct dynamic driving scenes, displaying high-quality reconstruction and real-time rendering.


\section{SLAM}\label{sec:slam}

Due to the powerful ability of NeRF to render images based on the pose and view orientation, the attempt to combine NeRF with pose estimation, as well as SLAM, is naturally considered and investigated by numerous researchers. Related research can be generally divided into two categories: pose estimation by NeRF and scene representation by NeRF. 



\subsection{Pose Estimation by NeRF}

Several specific approaches of estimating real-time pose by NeRF have been emerged recently, and can be categoriesd into 3D implicit representation and 3D feature extraction.

\subsubsection{3D Implicit Representation}

\begin{figure}
    \centering{\includegraphics[width = \linewidth]{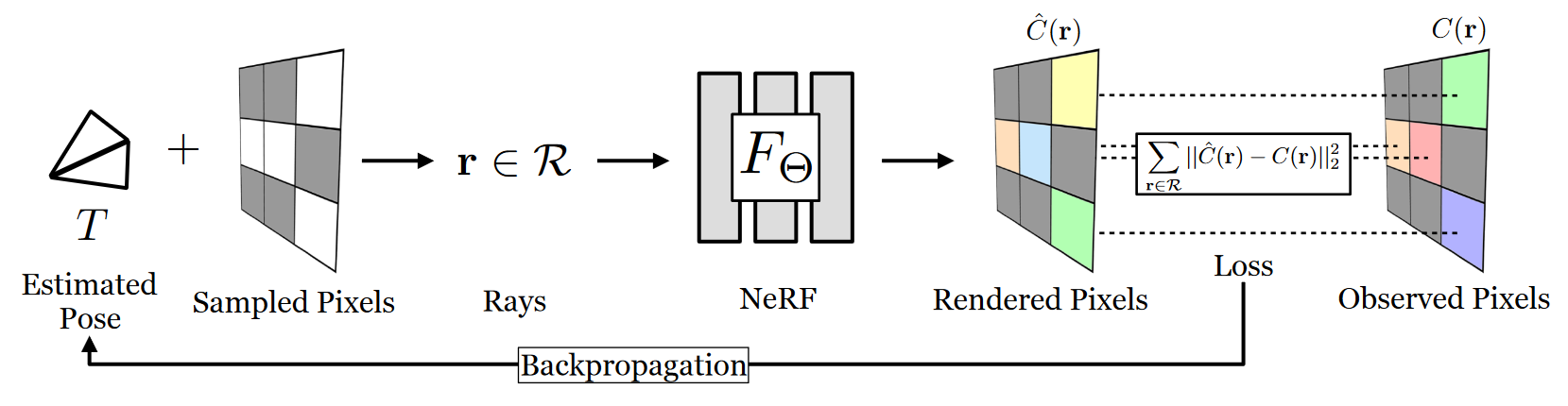}}
    \caption{The "inversion-like" pipeline of iNeRF \cite{yen2021inerf}}
    \label{fig:iNeRF}
\end{figure}

The most straightforward idea is to utilize the 3D implicit representation ability of NeRF to conduct relocalization\cite{yen2021inerf, adamkiewicz2022vision, maggio2023loc, katragadda2023nerf, kuang2023ir, moreau2022lens, hou2022implicit}. Considering the pipeline of NeRF, iNeRF\cite{yen2021inerf} presents an “inverting” pipeline to optimize pose estimation by a pre-trained NeRF as Fig. \ref{fig:iNeRF} shows. Rendered pixels are generated by NeRF from an estimated pose, which is then optimized by back-propagation of the residual between rendered and observed pixels. NeRF-Navigation\cite{adamkiewicz2022vision} further combines a process loss based on the dynamic model with the photometric loss to filter the tracking results and avoid pose initialization. 
Besides directly comparing the observed image with the rendered one, NeRF-VINS\cite{katragadda2023nerf} matches the observed image with the image generated by NeRF from the pose with a small offset to current estimated pose to update pose estimation. It is claimed by the authors that the synthetic image should have a significant overlapping field of view (FOV) with the observed image, which benefits the matching and pose estimation. A 2D LiDAR-based indoor Monte Carlo localization method is presented in IR-MCL\cite{kuang2023ir}, which predicts the occupancy probability for localization by neural network instead of volume density like NeRF. As the input and output are both light weight, IR-MCL realizes impressive real-time performance and generalizability. Considering NeRF's brilliant ability of novel view synthesis, LENS\cite{moreau2022lens} applies NeRF to augment the training dataset of a learning-based pose regressor, which is then used for real-time localization. Similarly, IMA\cite{hou2022implicit} trains a NeRF model conditioned on the sparse reconstruction generated by structure from motion(SfM), which is then densified by the trained NeRF to enhance relocalization.

\subsubsection{3D Feature Extraction}

However, above methods all require a well-trained NeRF right in the scenario. Several researchers regard NeRF as a well-generalized 3D feature extractor for different scenarios. NeRF-Loc \cite{liu2023nerf} designs a generalizable NeRF which is only conditioned on several supported images and depths to generate 3D descriptors from sampled 3D points. 2D descriptors are extracted from the query image to obtain 3D-2D correspondences and estimate relative pose by PnP (Perspective-n-Point) in a coarse-to-fine manner. In the meanwhile, Nerfels\cite{avraham2022nerfels} also notice the 3D representation capability of NeRF. Instead of overfitting a model to the entire scene, Nerfels represents scene-agnostic local 3D patches with renderable codes, improving the generalizability. A joint PnP+photometric optimization is performed in Nerfels, resulting in improvement of wide baseline pose estimation for both hand-crafted and learned local features.

\subsection{Scene representation by NeRF}
On the contrary of optimizing pose estimation by NeRF, another application of NeRF in SLAM is representing the whole scene to optimizing mapping performance. Based on the scene representation level, we classify related research into MLP-level, voxel-level, point-level and 3D Gaussian-level representation.

\subsubsection{MLP-level}
The idea of optimizing mapping performance by NeRF in SLAM is initially explored in iMAP\cite{sucar2021imap}, which establishes parallel tracking and mapping processes sharing one MLP as the scene representation and the same loss. The tracking process optimizes the pose with respect to the fixed scene network, just similar with the pipeline of iNeRF. While in the mapping process, after keyframe selection based on information gain and active sampling guided by render loss, the whole differentiable framework can be back-propagated to jointly optimize tracking and mapping performance. iMODE\cite{matsuki2023imode} further realizes large-scale incremental mapping without depth input. To extract more detailed features, Li et al.\cite{li2023end} propose a multi-MLP neural implicit coding structure.


\subsubsection{Voxel-level}
Rooted from traditional MLP-based NeRF, Instant-NGP\cite{M_ller_2022} encodes the scene into multi-resolution hash voxel vertices to realize real-time reconstruction, which enlights a group of NeRF-based SLAM research to represent the scene at voxel-level. The initial approach is proposed in Orbeez-SLAM\cite{chung2023orbeez}, which applies Instant-NGP in dense mapping based on the pose estimation and keyframe selection results from classic monocular SLAM algorithm. Successive research NGEL-SLAM\cite{mao2023ngel} include loop closure and global Bundle Adjustment(BA) for global pose refinement. Nevertheless, the aforementioned researches basically just integrate a mature SLAM system like ORB-SLAM2\cite{mur2017orb} into the NeRF framework, exhibiting no significant intrinsic innovation within NeRF itself.

Some other research fuses NeRF with SLAM in a closer manner, optimizing pose estimation performance by a rendered voxel-level NeRF\cite{zhu2022nice, yang2022vox, wang2023co, johari2023eslam, tang2023mips, zhu2023sni}. The most widely known research NICE-SLAM \cite{zhu2022nice} incorporates multi-level local information by a hierarchical feature voxel grids, which allow updates of coarse, mid and fine level local maps, while traditional single MLP is limited by scalability. The parallel tracking and mapping processes update alternatively as in iMAP. Furthermore, Vox-Fusion\cite{yang2022vox} incrementally allocates voxels by an octree-based structure without a pre-trained geometry decoder and proposes a keyframe selection strategy suitable for sparse voxels, resulting in better performance than NICE-SLAM on the Replica dataset in both tracking and mapping. Considering the memory footprint of voxel representation, ESLAM\cite{johari2023eslam} employs coarse-to-fine axis-aligned feature planes instead of feature voxel grids to reduce footprint growth by dimensionality reduction, while the divide-and-conquer scheme is leveraged in MIPS-Fusion\cite{tang2023mips} to realize scalable and robust SLAM incrementally by multi-implicit-submaps.  

Another exploration of NeRF in SLAM concentrates on large-scale mapping\cite{matsuki2023newton, haghighi2023neural, xiang2023nisb, liu2023efficient, deng2023plgslam}. Although Instant-NGP improves the reconstruction efficiency, it is still not capable for large-scale mapping because of the exponentially increasing of the hash table size. Therefore, NEWTON\cite{matsuki2023newton} firstly proposes a view-centric mapping approach with multiple local neural fields which are defined in local coordinate systems of each keyframe and dynamically allocated during the pose updating. Comparing with traditional world-centric neural field-based SLAM, NEWTON performs better in large-scale on-the-fly mapping. The large-scale scene is divided into multiple fix-sized cubes in \cite{haghighi2023neural, xiang2023nisb} to save computation costs. Additionally, Liu et al.\cite{liu2023efficient} focus on multi-agent implicit SLAM and propose a floating-point sparse voxel octree, based on which the local map points transforming can be realized by only adjusting three vertices of the octree, therefore significantly accelerates the map fusion between multi-agents.

While many researchers concern RGB-D images as input, or directly import depth estimation results from tracking, some researchers concentrate on handling depth uncertainty of monocular SLAM in mapping\cite{rosinol2022nerf, zhang2023go, hua2023fmapping, zhang2023hi, li2023dense, zhu2023nicer, sandstrom2023uncle}. In NeRF-SLAM \cite{rosinol2022nerf}, an uncertainty-based depth loss function is firstly designed to fully utilize SLAM outputs:

\begin{equation}
    \mathcal{L}_{D}(\boldsymbol{T}, \Theta) = \left\| D - D^{*}(\boldsymbol{T}, \Theta) \right\|^{2}_{\Sigma_D},
\end{equation}
where $D^{*}$ is the rendered depth, and $D, \Sigma_D$ are the input dense depth and the depth uncertainty estimated in the tracking procedure. FMapping\cite{hua2023fmapping} conducts a thorough theoretical analysis to examine the depth uncertainty, and divides the uncertainty into the initialization stage and on-the-fly mapping stage, which are then managed by factorized radiance field and sliding window sampling, respectively. Another approach to cope with depth uncertainty is proposed in HI-SLAM\cite{zhang2023hi} in the ray sampling procedure by sampling pixels with lower depth variance more frequently. Concentrates on the depth uncertainty of RGB-D inputs, UncLe-SLAM\cite{sandstrom2023uncle} import a Laplacian noise distribution of depth independently in each pixel.

\begin{figure}
    \centering{\includegraphics[width = \linewidth]{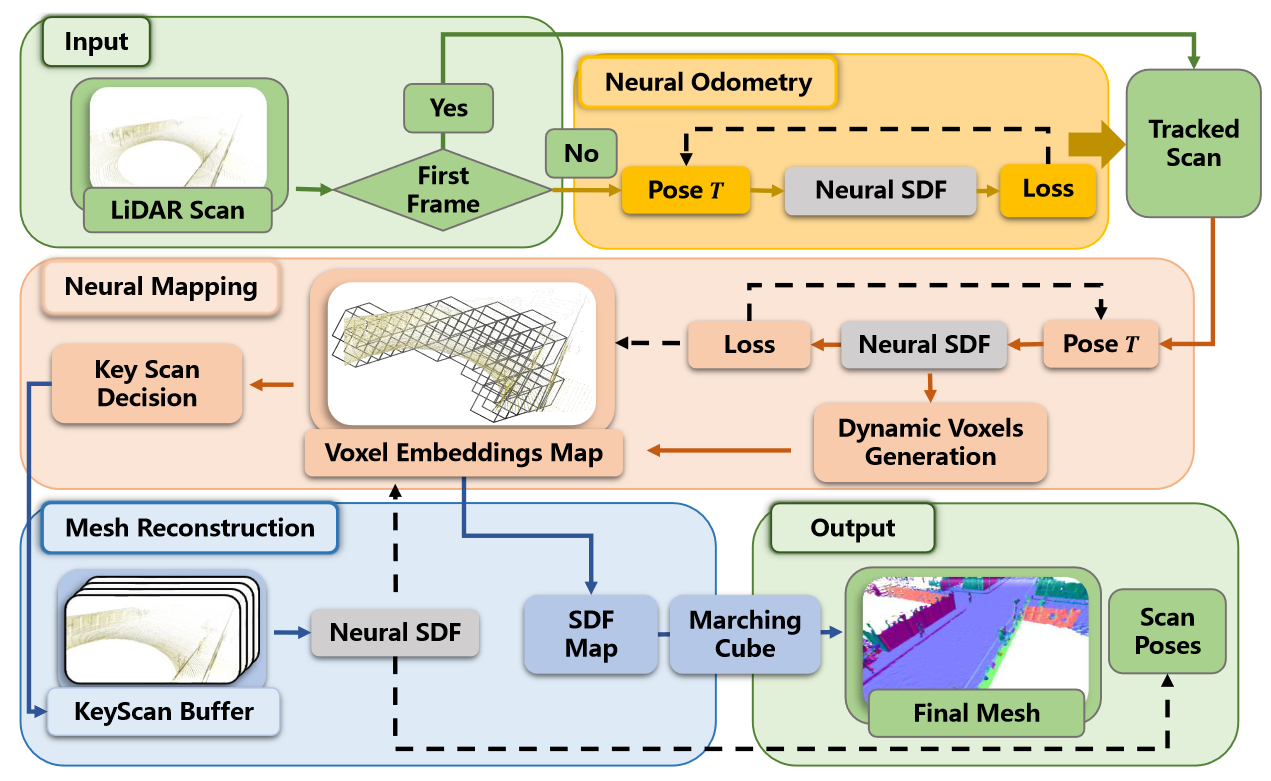}}
    \caption{The architecture of NeRF-LOAM \cite{deng2023nerf}}
    \label{fig:NeRF-LOAM}
\end{figure}

Apart from RGB/RGB-D as inputs, plenty of researchers also engage with other modalities as inputs\cite{deng2023nerf, isaacson2023loner, liu2023multi}. NeRF-LOAM\cite{deng2023nerf} formulates a neural signed distance function(SDF) tailored for LiDAR data, which can be used to conduct tracking, mapping and key-scan selection simultaneously. The architecture is shown in Fig .\ref{fig:NeRF-LOAM}. However, due to the time-consuming intersection query between the ray and the map, NeRF-LOAM is not able to operate in real time. LONER\cite{isaacson2023loner} proposes a loss function derived from Jensen-Shannon(JS) Divergence to accelerate convergence while refining the real-time reconstruction performance.

Some other remarkable scene representations in NeRF-based SLAM have arisen recently. Teigen et al.\cite{teigen2023rgb} design a plenoxel radiance field-based SLAM system motivated by the plenoxel\cite{fridovich2022plenoxels}, which is an analytical radiance field representation without neural network but rather a voxel grid representation. The analytical derivative equations for tracking and mapping show both result improvement and time reduction compared with neural network-based methods.  NeRF-based SLAM is explored in structural environments in Structerf-SLAM\cite{wang2024structerf} by structured planar constraints. 


\subsubsection{Point-level}
Neural point cloud-based scene representations are also anticipated to be capable for large-scale real-time tracking and mapping, as the structure of point cloud is not so compact as grids and is suitable for dynamic allocation. Point-SLAM\cite{sandstrom2023point} executes this strategy and dynamically adapts the anchor point density to the information density of the input RGBD image, rendering different levels of detail with different point densities, thus achieving competitive results to other dense neural RGBD SLAM methods in both tracking and mapping. 

As point cloud-based representation is more light-weight and appropriate for loop closure and global pose graph optimization than MLP or voxel, CP-SLAM\cite{hu2023cp} facilitates the multi-agent SLAM system with loop closure for a single agent and cooperative localization and mapping for multiple agents with the advantage of neural point cloud. Similarly, Loopy-SLAM\cite{liso2024loopy} designs point cloud submaps that grow iteratively to perform loop closure and reduce error accumulation. PIN-SLAM\cite{pan2024pin} also incorporates point-based implicit neural representation to achieve large-scale SLAM by LiDAR.


\subsubsection{3D Gaussian-level}

With the rapid development of the recent 3D Gaussian Splatting\cite{kerbl20233d}, plenty of 3D Gaussian-level SLAM are emerging. The first group of this type SLAM \cite{yan2023gs, matsuki2023gaussian, yugay2023gaussian,keetha2023splatam} integrate explicit 3D Gaussian representation to boost both tracking and mapping performance benefiting from the fast splatting rendering technique. Recently, SemGauss-SLAM\cite{zhu2024semgauss} and SGS-SLAM\cite{li2024sgs} further incorporate semantic information to guide bundle adjustment and construct semantic maps for downstream tasks.

\section{Simulation}\label{sec:simulation}
Autonomous driving simulation offers a safer and more cost-effective alternative to real-world testing by creating realistic virtual environments for sensor data generation, which facilitates the creation of diverse driving scenarios and reduces safety risks.  Traditional simulation methods like CARLA \cite{dosovitskiy2017carla} and AirSim \cite{shah2017airsim}, which rely on manual scene creation and have a significant sim-to-real gap due to handcrafted assets and simplified physics, face limitations. GeoSim \cite{chen2021geosim} attempts to bridge this gap by combining graphics and neural networks for video scene generation but fails to simulate sensor data for new views.  The Neural Radiance Field approach significantly enhances realism and reduces manual effort in scene creation and editing, presenting a promising solution to narrow the domain gap between the real and virtual worlds. Methods for simulation fall into two main categories: image data simulation and LiDAR data simulation. We will discuss them in the following parts.

\subsection{Image Data Simulation}
The current image data simulation methods for autonomous driving based on Neural Radiance Fields involve reconstructing scenes by using a sequence of images from real driving environments along with corresponding camera poses, allowing for the modification of vehicle behaviors within the original scenes to generate and render new photorealistic images. 
\begin{figure}[htbp]
    \centering
    \includegraphics[width=0.5\textwidth]{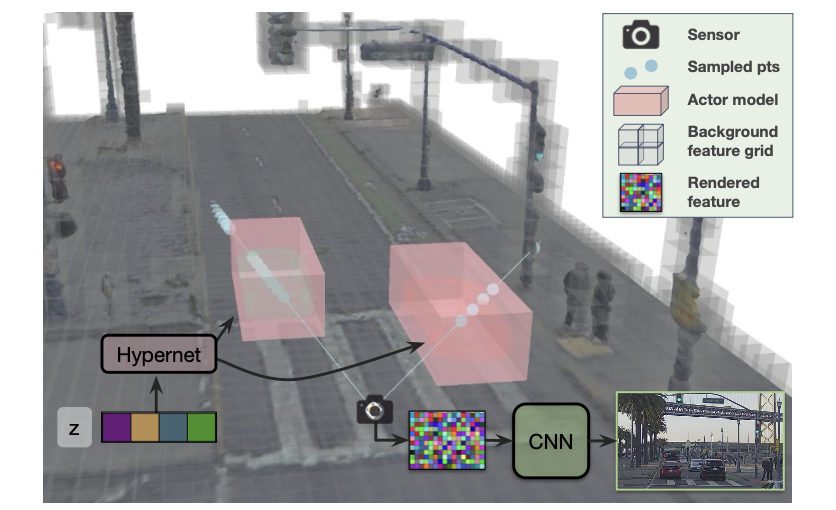}
    \caption{Overview of UniSim\cite{yang2023unisim}: First, divide the 3D scene into a static background (grey) and a set of dynamic actors (red).}
    \label{unisim_pip}
\end{figure}
Depending on the representation technique, these methods are further categorized into implicit representation approaches, exemplified by NeRF, and explicit representation approaches, represented by 3D Gaussian Splatting \cite{kerbl20233d}.

\subsubsection{Implicit Representation}
\begin{figure*}[ht]
  \includegraphics[width=\textwidth,height=3cm]{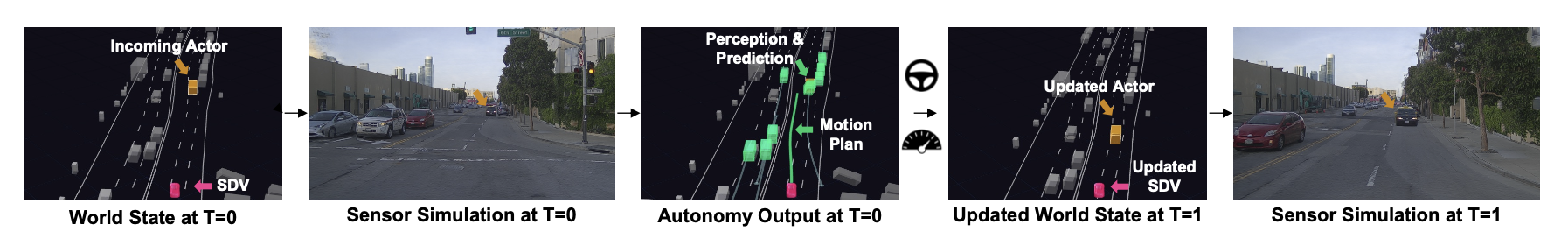}
  \caption{UniSim Close-loop Autonomy Evaluation}
  \label{Closed-loop Evaluation}
\end{figure*}
These methods utilize implicit representation models similar to NeRF to reconstruct scenes. NSG\cite{ost2021neural} employs a vanilla NeRF model to represent the static background. For vehicle reconstruction, NSG reconstructs vehicles of the same category with a NeRF model, assigning each vehicle a latent code for appearance reconstruction. After training, NSG can edit the pose of vehicles in the scene by controlling their 3D bounding boxes, generating new scenes, and ultimately rendering photorealistic images. NSG transforms complex dynamic scene tasks into 3D reconstructions of multiple independent static objects by decomposing the scene into static backgrounds and vehicles using 3D bounding boxes. Limited by the capabilities of the vanilla NeRF model, NSG suffers from long training times and poor rendering quality. Instant-NGP \cite{M_ller_2022} improves the efficiency of scene reconstruction and image rendering quality through the use of multilevel hash grid encoding. UniSim \cite{yang2023unisim} adopts NSG's method of scene decomposition, using separate Instant-NGP models to reconstruct static backgrounds and vehicles respectively, as shown in Fig .\ref{unisim_pip}. To further improve the efficiency of background reconstruction, UniSim utilizes geometry priors from LiDAR observations to identify near-surface voxels and optimize only their features. When dealing with vehicles in the scene, UniSim uses a hypernetwork \cite{ha2016hypernetworks} to generate the representation of each vehicle from a learnable latent. UniSim employs Closed-loop Evaluation to demonstrate that their simulation data can be used to test the performance of autonomous vehicles in safety-critical scenarios (Figure .\ref{Closed-loop Evaluation}). 

As NeRF models are continuously optimized, the aforementioned methods are limited by the NeRF-related backbones used for scene representation. MARS\cite{wu2023mars} utilizes the framework of the NeRFStudio platform, which allows for flexible switching between different modern NeRF-related backbones and sampling strategies, to design a modular model. Moreover, besides rendering RGB images, MARS can also generate semantic segmentation images and depth maps of the scene.
However, these methods require multiple neural radiance fields to represent elements in the scene, reducing the efficiency of scene reconstruction. NeuRAD \cite{tonderski2023neurad} encodes the static background and vehicles in the scene through different multilevel hash grids and reconstructs them together using a shared neural radiance field.

\subsubsection{Explicit Representation}
\begin{figure}[htbp]
    \centering
    \includegraphics[width=\linewidth]{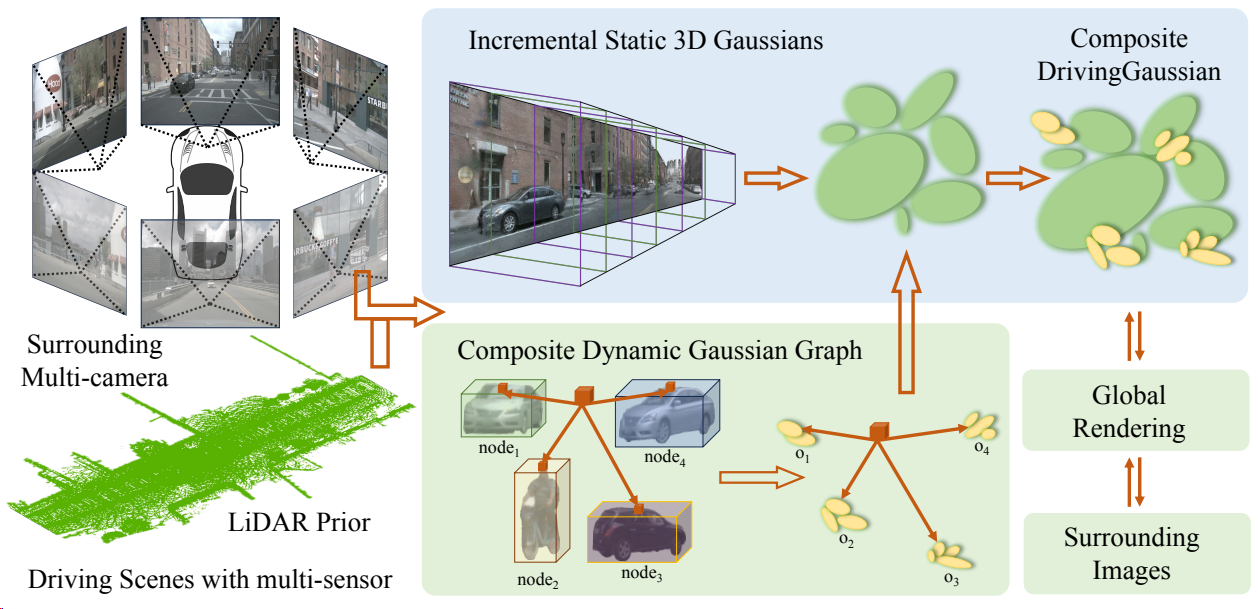}
    \caption{Overview of DrivingGaussian.}
    \label{fig:DrivingGaussian}
\end{figure}

Due to the frequent use of MLP to query information about points in a scene, their training and rendering times cannot meet real-time requirements. 3D Gaussian Splatting \cite{kerbl20233d} (3DGS) can generate high-quality new viewpoint images and scene geometries while meeting real-time 3D reconstruction requirements, making simulation methods based on 3DGS for autonomous driving increasingly popular. 
DrivingGaussian \cite{zhou2024drivinggaussian} initializes the positions of 3D gaussians using LiDAR point cloud data and, similar to other methods, employs 3D bounding boxes to decompose the scene's 3D Gaussians into static backgrounds and vehicles within the scene, as shown in Figure \ref{fig:DrivingGaussian}. To apply 3DGS to large-scale static backgrounds, DrivingGaussian enhances 3DGS by introducing Incremental Static 3D Gaussians, reconstructing the complete static background by decomposing the background into multiple independent small bins and sequentially initializing the positions of 3D gaussians in each bin. 
DrivingGaussian uses 3D bounding boxes provided by the nuScenes dataset as the true positions of all vehicles, but in practical applications, a tracker model \cite{wu2022transformationequivariant} is needed to predict the 3D bounding boxes of vehicles in images. However, tracker model-generated bounding boxes are generally noisy. Directly using them to optimize scene representation leads to a decrease in rendering quality. To address this issue, Street Gaussians \cite{yan2024street} treats tracked poses as learnable parameters by adding a learnable transformation to each vehicle's transformation matrix. Unlike other methods based on 3DGS, Street Gaussians employ a 4D Spherical Harmonics (4D SH) model for reconstructing the color of dynamic vehicles. This allows vehicles to exhibit appearances that change over time. Street Gaussians also assigns a semantic parameter to each 3D Gaussian in space through supervised learning, aiding the model's understanding of 3D scenes. Furthermore, to achieve holistic 3D scene understanding, HUGS \cite{zhou2024hugs} also predicts the optical flow information of the scene, in addition to its RGB images and semantic information. Additionally, HUGS uses physical constraints derived from the unicycle model to optimize the trajectory of each vehicle's 3D bounding box, resulting in more accurate and smooth trajectories.

\subsection{LiDAR Data Simulation}
The aim of LiDAR data simulation is to utilize LiDAR measurement data to enhance neural scene representation, thus facilitating the synthesis of realistic LiDAR scans from novel viewpoints. Grounded in distinct LiDAR sensing process modeling techniques, these methodologies are mainly divided into two classifications: ray models and beam models. The following text will introduce these two methods respectively.
\subsubsection{Ray Model}
These methods simplify the LiDAR sensing process into a single ray, replacing the camera ray in the original NeRF model, and transform the LiDAR point cloud data into 360-degree panoramic images through spherical projection as ground truth, converting point cloud data into pseudo image data. NeRF-LiDAR \cite{zhang2023nerflidar} uses LiDAR point cloud data with semantic labels as ground truth, reconstructing 3D scenes through a neural radiance field and generating LiDAR point clouds with accurate semantic labels. To accurately reproduce the LiDAR ray dropping phenomenon, NeRF-LiDAR predicts the locations where this phenomenon occurs by training a classification mask on panoramic images. Although NeRF-LiDAR can generate LiDAR point cloud data with semantic information, it does not predict the important data of ray intensity. LiDAR-NeRF \cite{tao2023lidar} also converts LiDAR point cloud data into panoramic images and produces 3D representations of the distance, the intensity, and the ray dropping probability at each pseudo-pixel. NeRF-like methods display inferior geometry in low-texture areas of large-scale scenes. To overcome this limitation, LiDAR-NeRF incorporates a structural regularization to preserve local structural details, thereby improving NeRF's ability to reconstruct geometric shapes more effectively.
\subsubsection{Beam Model}
\begin{figure*}[ht]
  \includegraphics[width=\textwidth]{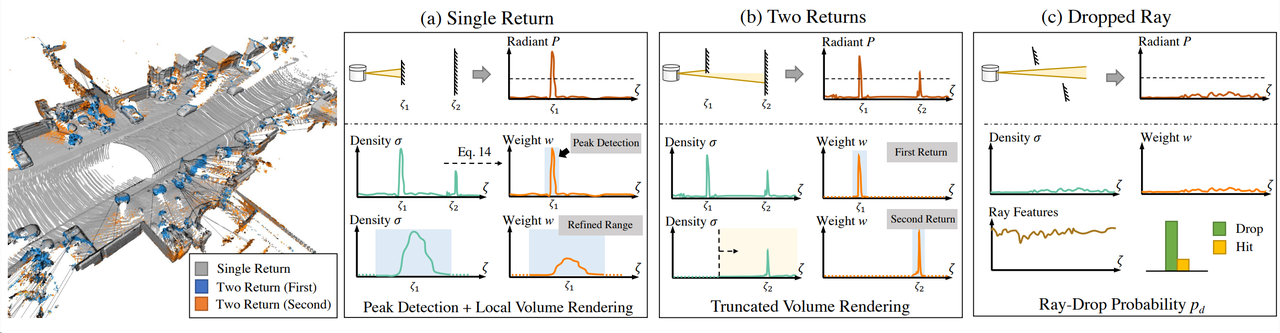}
  \caption{The overall pipeline of NFL. (a) Single Return, (b) Two Returns, (c) Dropped Ray.}
  \label{fig:NFL}
\end{figure*}

Unlike the aforementioned methods, NFL \cite{huang2023neural} uses diverged beams with scattering angles to simulate the LiDAR sensing process. This technique can accurately reproduce key sensor behaviors such as beam divergence, secondary returns, and ray dropping, as shown in Figure \ref{fig:NFL}. 

\section{Discussion}\label{sec:discuss}

\subsection{Perception}
In the context of the data branch, NeRF has garnered attention for its potential applications. While neural radiance fields (NeRF) have been explored for generating single-frame images, this approach proves inadequate for algorithms that depend on multi-frame inputs. Various studies, including BEVFormer\cite{bevformer} and the Sparse4D series\cite{sparse4d, sparse4dv2, sparse4dv3}, have showcased the effectiveness of integrating temporal information. The exploration of NeRF's ability to unlock the temporal-consistent data augmentation is urgently needed.

In the context of the model branch, NeRF utilizes an implicit scene representation and neural rendering to connect 3D scenes with 2D images, demonstrating advancements in diverse perception tasks. However, the computational inefficiency of the neural rendering process poses a challenge in effectively sampling rays while upholding multi-view and temporal consistency, particularly in high-dynamic scenarios like autonomous driving.

\subsection{Reconstruction}
In the reconstruction domain, several applications have been adapted to solve industrial problems, such as transforming the data collected in one sensor setup to another sensor setup (mostly the camera intrinsic and extrinsic) to adapt to new cars, and create new data for augmentation and evaluation. However, most of the methods are limited to reconstructing rigid dynamic scenes, such as static streets with only moving vehicles, and cannot handle non-rigid dynamic objects like walking pedestrians. Future work can incorporate more object-prior to reconstruct objects. In the meantime, the reconstruction quality and runtime can still be a limitation of current methods. Future work can leverage generalizable-prior such as NeRFusion~\cite{zhang2022nerfusion} to speed up reconstruction.

Furthermore, potential improvement can be leveraging the recent advances in Generative AI, generating an unlimited amount of data that is not restricted to only reconstructing real-world data. For example, researchers can use Sora from OpenAI to first generate realistic video, then reconstruct it to 3D representation, to enable diverse 3D generation.

\subsection{SLAM}
Existing NeRF-based SLAM research has the ability for autonomous driving localization and mapping. Moreover, ground truth auto-labeling and online extrinsic calibration are two potential fields for NeRF-based SLAM research.

However, current research concentrates mostly on indoor scenarios, which, though the techniques can be referred to in autonomous driving, are still not so capable of handling outdoor large-scale scenarios. Furthermore, the dynamic characteristic in autonomous driving greatly impact traditional NeRF-based SLAM research as they are tend to suffer from time-varying scenarios. To enable NeRF-based SLAM for autonomous driving, a more light-weight data structure for mapping in large-scale scenarios is in urgent need. Besides, strategies to reduce the influence of dynamic objects are also a necessity. 

Another non-negligible factor is the light condition. In autonomous driving, a great number of scenarios contain severe light condition such as night or abnormal weathers like snow and fog. How to improve the robustness of NeRF-based SLAM in these scenarios presents a huge challenge. One possible solution is to introduce robust sensors such as radars to serve as a supplement.

\subsection{Simulation}
Current simulation techniques based on Neural Radiance Fields rely on a multitude of images captured from multiple perspectives to achieve more accurate geometric restoration when reconstructing vast urban scenes. Neither NeRF nor 3D Gaussian Splatting techniques can precisely restore scenes within visual blind spots, primarily due to the insufficient generalization capability of these models during scene reconstruction, failing to recover complete overall scenes from limited sparse viewpoint data. Therefore, future work will require methods based on few-shot view synthesis to address the accurate reconstruction of scenes with limited views.

Secondly, the lack of realistic interactive feedback between vehicle appearances and scene lighting leads to compromised authenticity of rendered images. Existing methods reconstruct vehicles and scenes as independent elements, thereby ignoring the impact of scene lighting on vehicle appearance. In the future, appearance editing for objects could be integrated with traditional computer graphics shading algorithms. 

Moreover, objects reconstructed by existing methods are rigid, their geometric shapes do not change over time, making it a challenge to reconstruct and edit deformable objects like pedestrians while reconstructing the entire scene. In the next phase of research, it might be possible to combine existing deformable human model reconstruction methods to achieve reconstruction and editing capabilities for pedestrians.

\section{Conclusion}\label{sec:conclu}
In this survey, we give a comprehensive review of neural radiance field in the context of Autonomous Driving (AD). To be specific, we first introduce the basic principles and background of NeRF, then delve into a comprehensive analysis of the application of NeRF in various fields of AD, categorized as Perception, 3D Reconstruction, SLAM, and Simulation. At last, we discuss the remaining challenges in each category and provide possible solutions. We hope this survey will facilitate future research work and promote the arrival of the era of autonomous driving.

\bibliographystyle{IEEEtran}
\bibliography{ref}

\begin{IEEEbiography}
[{\includegraphics[width=1in,height=1.25in,clip,keepaspectratio]{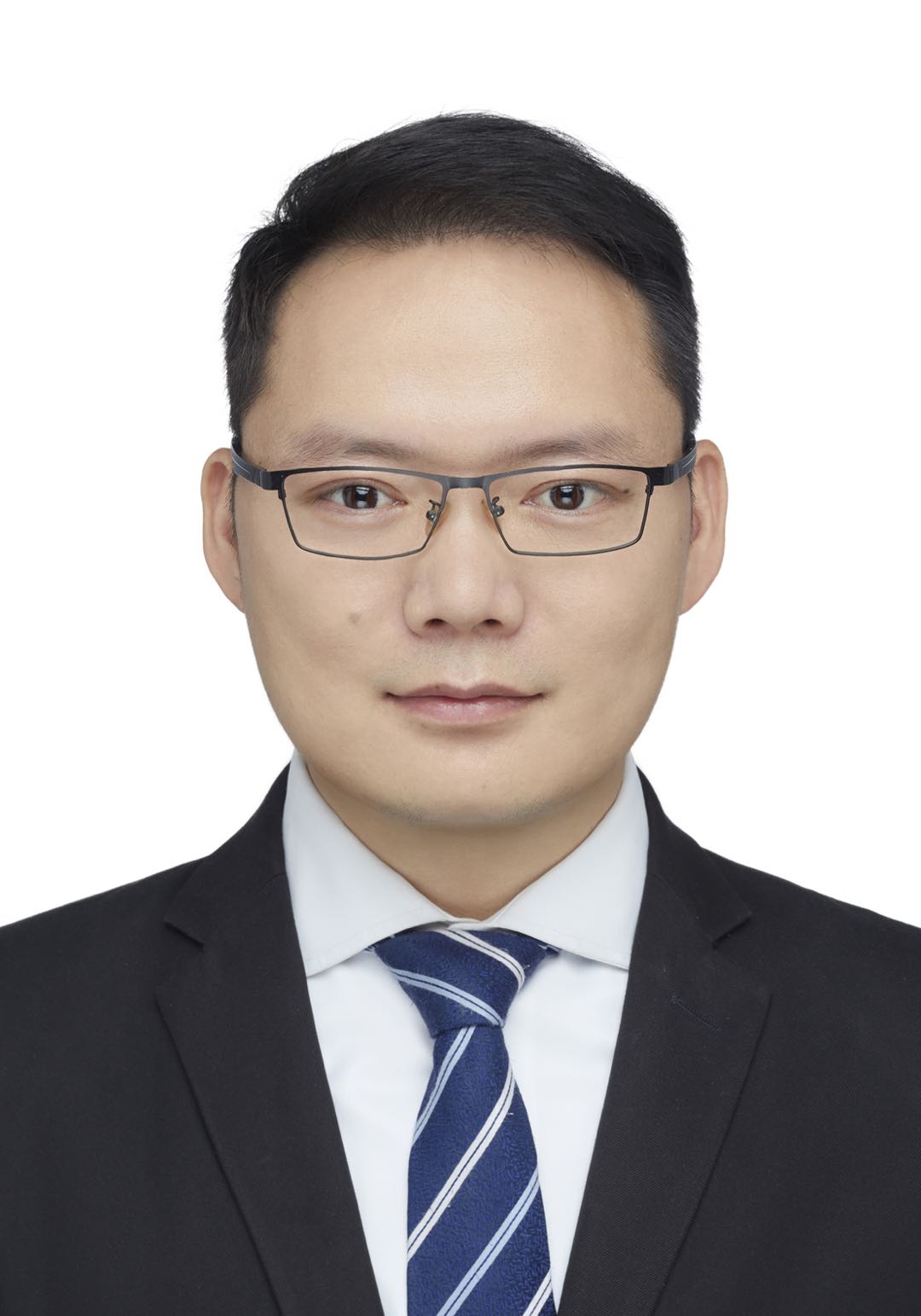}}]
{Lei He} 
(helei2023@tsinghua.edu.cn) received his B.S. in Beijing University of Aeronautics and Astronautics, China, in 2013, and the Ph.D. in the National Laboratory of Pattern Recognition, Chinese Academy of Sciences, in 2018. From then to 2021, Dr. He served as a postdoctoral fellow in the Department of Automation, Tsinghua University, Beijing, China. He worked as the research leader of the Autonomous Driving algorithm at Baidu and NIO from 2018 to 2023. He is a Research Scientist in automotive engineering with Tsinghua University. His research interests include Perception, SLAM, Planning, and Control.
\end{IEEEbiography}

\begin{IEEEbiography}
[{\includegraphics[width=1in,height=1.25in,clip,keepaspectratio]{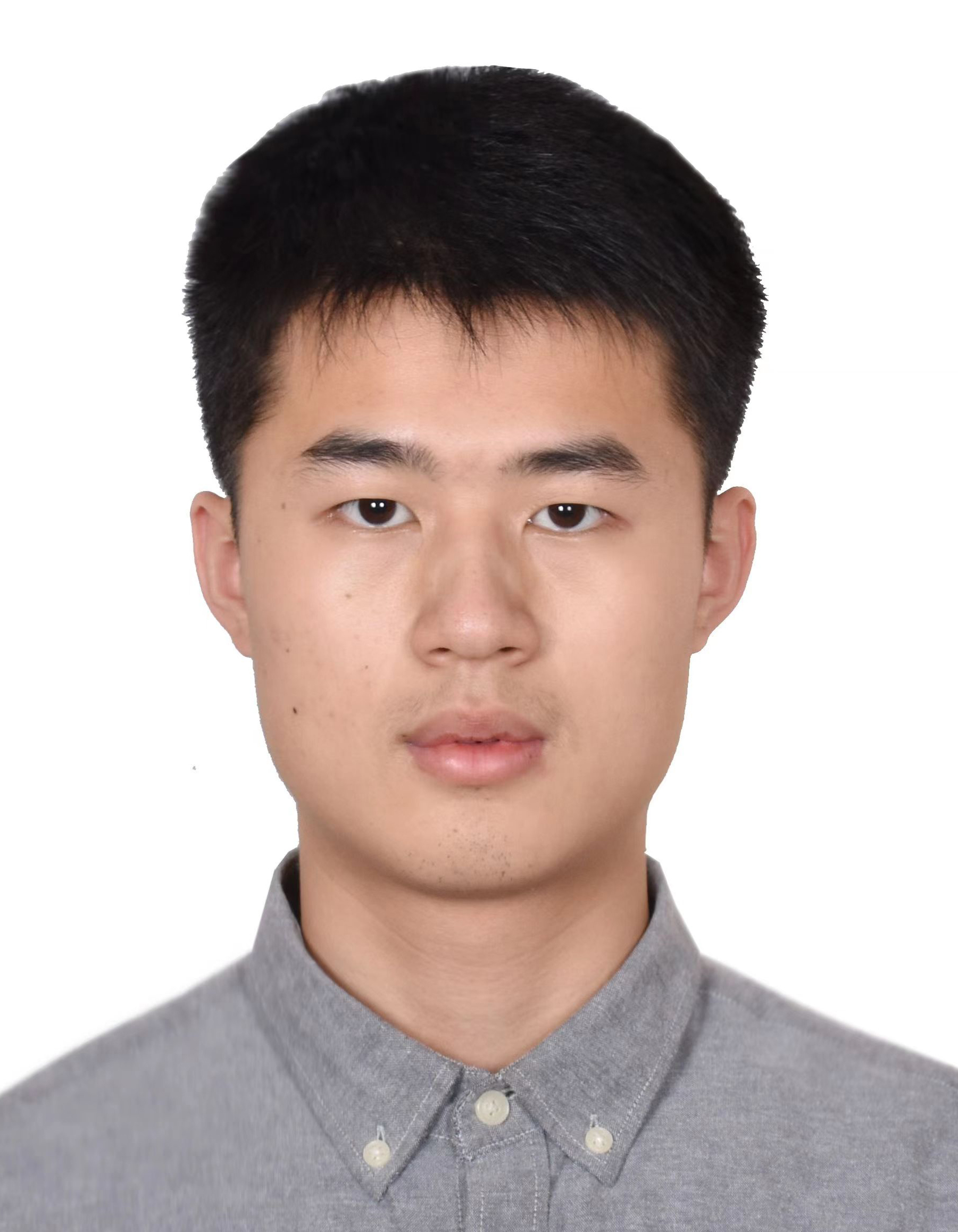}}]
{Leheng Li} 
(lli181@connect.hkust-gz.edu.cn) earned his bachelor's degree in mathematics from the School of Mathematical Sciences, Dalian University of Technology, Dalian, China, in 2022. He is currently a Ph.D student in Artificial Intelligence at Information Hub from The Hong Kong University of Science and Technology (Guangzhou), Guangzhou 510000, China. His research insterests are computer vision and autonomous driving.
\end{IEEEbiography}

\begin{IEEEbiography}
[{\includegraphics[width=1in,height=1.25in,clip,keepaspectratio]{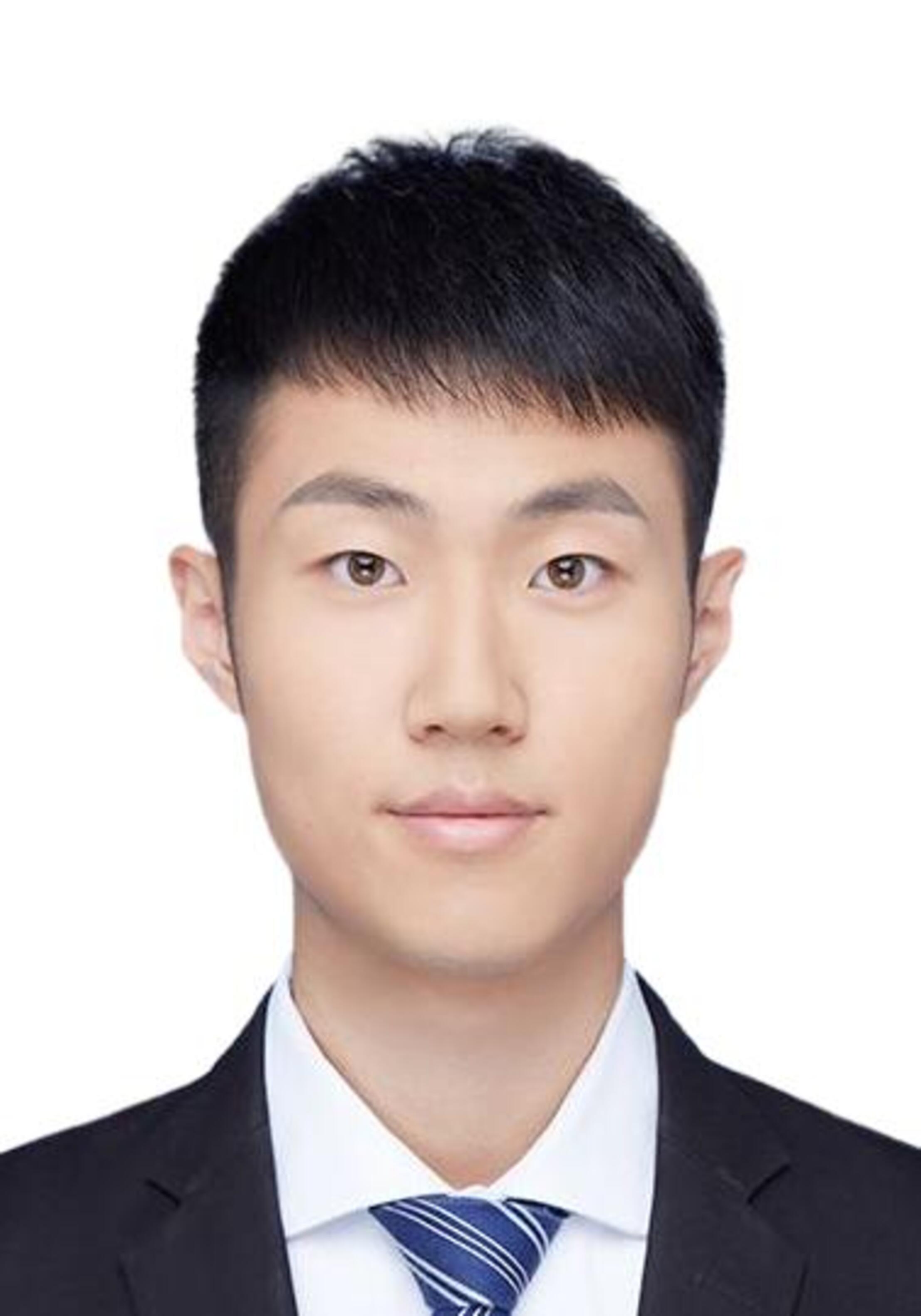}}]
{Wenchao Sun}
(swc21@mails.tsinghua.edu.cn) received the B.E. degree from Tsinghua University, Beijing, China, in 2021. He is currently a Ph.D. student in mechanical engineering
with the School of Vehicle and Mobility, Tsinghua University. His research interests include end-to-end autonomous driving and simulation.
\end{IEEEbiography}

\begin{IEEEbiography}
[{\includegraphics[width=1in,height=1.25in,clip,keepaspectratio]{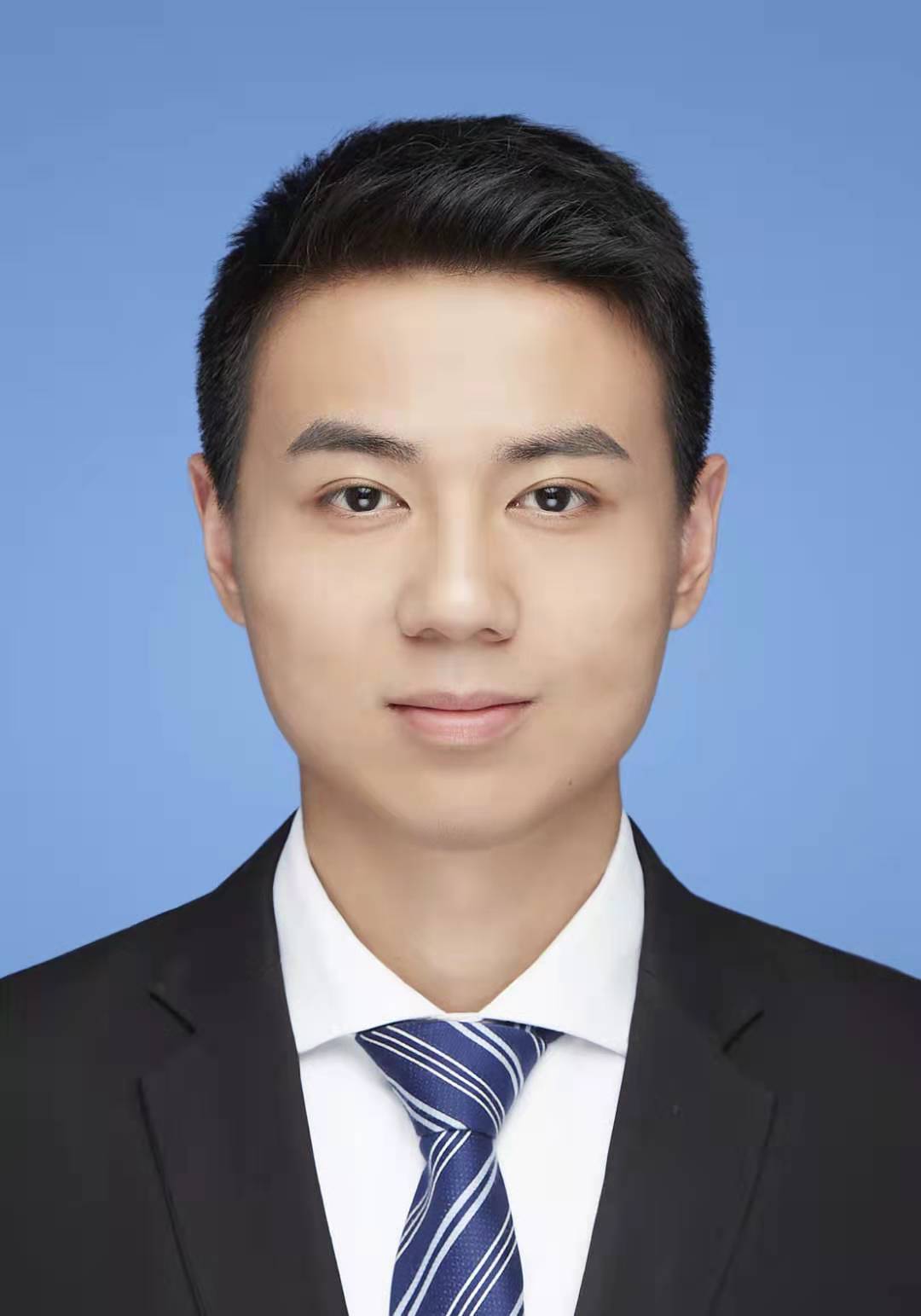}}]
{Zeyu Han} 
(hanzy21@mails.tsinghua.edu.cn) received the bachelor’s degree in automotive engineering from School of Vehicle and Mobility, Tsinghua University, Beijing, China, in 2021. He is currently pursuing the Ph.D. degree in mechanical engineering with School of Vehicle and Mobility, Tsinghua University, Beijing, China. His research interests includes autonomous driving SLAM and environment understanding by 4D mmWave radars.
\end{IEEEbiography}

\begin{IEEEbiography}
[{\includegraphics[width=1in,height=1.25in,clip,keepaspectratio]{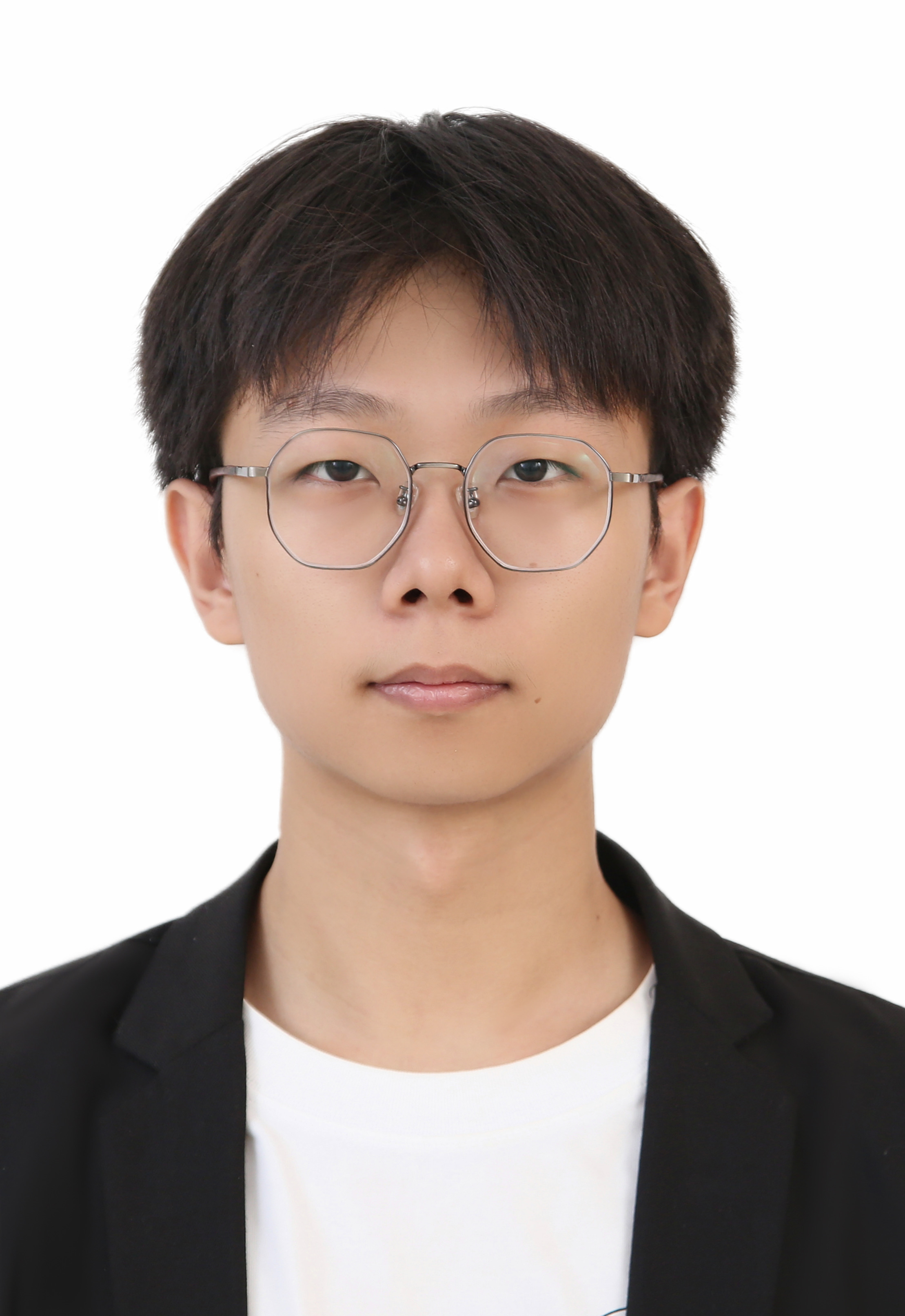}}]
{Yichen Liu} 
(nz23750@bristol.ac.uk) received his bachelor of engineering degree in Telecommunications Engineering with Management from Beijing University of Posts and Telecommunications, Beijing, in 2023. He is currently pursuing a Master's degree in Robotics at the University of Bristol, Bristol, UK. His research interests include computer vision and autonomous driving.
\end{IEEEbiography}

\begin{IEEEbiography}
[{\includegraphics[width=1in,height=1.25in,clip,keepaspectratio]{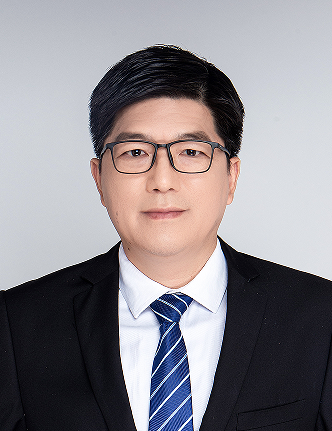}}]
{Sifa Zheng} 
(zsf@tsinghua.edu.cn) received the B.E. and Ph.D. degrees from Tsinghua University, Beijing, China, in 1993 and 1997, respectively. He is currently a professor in the School of Vehicle and Mobility, and the State Key Laboratory of Automotive Safety and Energy, Tsinghua University. He is also the deputy director, Suzhou Automotive Research Institute, Tsinghua University. His current research interests include autonomous driving, vehicle dynamics and control.
\end{IEEEbiography}

\begin{IEEEbiography}
[{\includegraphics[width=1in,height=1.25in,clip,keepaspectratio]{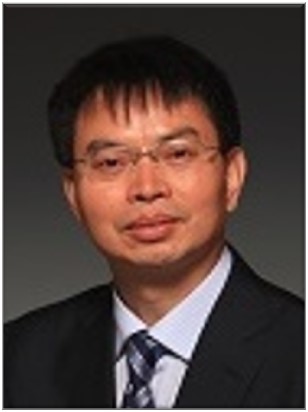}}]
{Jianqiang Wang} 
(wjqlws@tsinghua.edu.cn) received the B. Tech. and M.S. degrees from Jilin University of Technology, Changchun, China, in 1994 and 1997, respectively, and Ph.D. degree from Jilin University, Changchun, in 2002. He is currently a Professor of School of Vehicle and Mobility, Tsinghua University, Beijing, China. He has authored over 150 papers and is a co-inventor of over 140 patent applications. He was involved in over 10 sponsored projects. His active research interests include intelligent vehicles, driving assistance systems, and driver behavior. He was a recipient of the Best Paper Award in the 2014 IEEE Intelligent Vehicle Symposium, the Best Paper Award in the 14th ITS Asia Pacific Forum, the Best Paper Award in 2017 IEEE Intelligent Vehicle Symposium, the Changjiang Scholar Program Professor in 2017, Distinguished Young Scientists of NSF China in 2016, and New Century Excellent Talents in 2008.
\end{IEEEbiography}

\begin{IEEEbiography}
[{\includegraphics[width=1in,height=1.25in,clip,keepaspectratio]{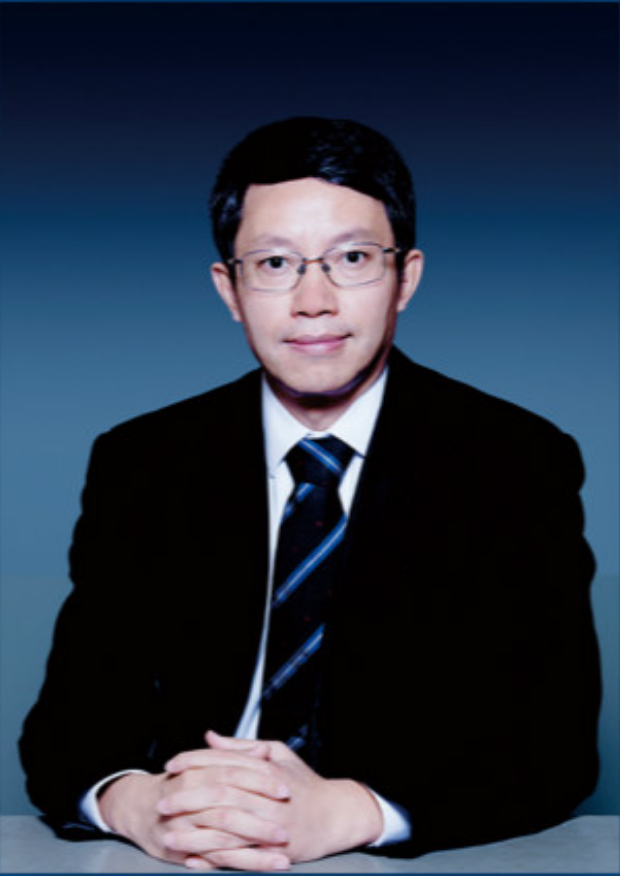}}]
{Keqiang Li} 
(likq@tsinghua.edu.cn) received the B.Tech. degree from Tsinghua University of China, Beijing, China, in 1985, and the M.S. and Ph.D. degrees in mechanical engineering from the Chongqing University of China, Chongqing, China, in 1988 and 1995, respectively. He is currently a Professor with the School of Vehicle and Mobility, Tsinghua University. His main research areas include automotive control system, driver assistance system, and networked dynamics and control, and is leading the national key project on CAVs (Connected and Automated Vehicles) in China. Dr. Li has authored more than 200 papers and is co-inventors of over 80 patents in China and Japan. Dr. Li has served as a Fellow Member of Chinese Academy of Engineering, a Fellow Member of Society of Automotive Engineers of China, editorial boards of the International Journal of Vehicle Autonomous Systems, Chairperson of Expert Committee of the China Industrial Technology Innovation Strategic Alliance for CAVs (CACAV), and CTO of China CAV Research Institute Company Ltd. (CCAV). He has been a recipient of Changjiang Scholar Program Professor, National Award for Technological Invention in China, etc.

\end{IEEEbiography}

\vfill

\end{document}